\documentclass[letterpaper, 10 pt, conference]{IEEEexample}  

\usepackage{algorithm} 
\usepackage{mathrsfs}
\usepackage{amsmath}
\usepackage{amsfonts}
\usepackage{hyperref}
\usepackage{url}
\usepackage{graphicx}
\usepackage{subcaption}

\usepackage{booktabs}
\usepackage{siunitx}
\usepackage{algpseudocode}
\usepackage{varwidth}

\usepackage{xcolor} 

\newcommand{\ijf}[1]{{\textcolor{black}{#1}}}

\IEEEoverridecommandlockouts                              

\title{\LARGE \bf
Continuous Transition: Improving Sample Efficiency for Continuous Control Problems via MixUp
}

\author{\ijf{Junfan Lin$^{1}$, Zhongzhan Huang$^{3}$, Keze Wang$^{3,4*}$, Xiaodan Liang$^{2,3}$, Weiwei Chen$^{3}$, and Liang Lin$^{1,3}$}
\thanks{\ijf{$^{1}$ Sun Yat-sen University, Guangzhou, China, 510006.}}
\thanks{\ijf{$^{2}$ Shenzhen Campus of Sun Yat-sen University, Shenzhen, China, 518000.}}
\thanks{\ijf{$^{3}$ Dark Matter AI Inc.}}
\thanks{\ijf{$^{4}$ University of California, Los Angeles.}}
\thanks{\ijf{$^{*}$ Corresponding author.}}
\thanks{\ijf{Email: {\tt\small linjf8@mail2.sysu.edu.cn}, {\tt\small zhongzhanhuang@foxmail.com}, {\tt\small \{kezewang, xdliang328\}@gmail.com}, {\tt\small benwwc@outlook.com, linliang@ieee.org}}}%
}

\begin{document}

\maketitle
\thispagestyle{empty}
\pagestyle{empty}

\begin{abstract}
Although deep reinforcement learning~(RL) has been successfully applied to a variety of robotic control tasks, it's still challenging to apply it to real-world tasks, due to the poor sample efficiency. Attempting to overcome this shortcoming, several works focus on reusing the collected trajectory data during the training by decomposing them into a set of policy-irrelevant discrete transitions. However, their improvements are somewhat marginal since i) the amount of the transitions is usually small, and ii) the value assignment only happens in the joint states. To address these issues, this paper introduces a concise yet powerful method to construct \textit{Continuous Transition}, which exploits the trajectory information by exploiting the potential transitions along the trajectory. Specifically, we propose to synthesize new transitions for training by linearly interpolating the \ijf{consecutive} transitions. To keep the constructed transitions authentic, we also develop a discriminator to guide the construction process automatically. 
Extensive experiments demonstrate that our proposed method achieves a significant improvement in sample efficiency on various complex continuous robotic control problems in MuJoCo and outperforms the advanced model-based / model-free RL methods. \ijf{The source code is available\footnote{\ijf{\href{https://github.com/junfanlin/continuous-transition}{github.com/junfanlin/continuous-transition}}}.}

\end{abstract}

\section{INTRODUCTION}

Deep reinforcement learning~(RL), with high-capacity deep neural networks, has been applied to solve various complex decision-making tasks, including chess games~\cite{silver2017mastering,silver2018general}, video games~\cite{mnih2015human,vinyals2019grandmaster}, etc. In robotics, the ultimate goal of RL is to endow robots with the ability to learn, improve, adapt, and reproduce tasks (e.g., robotic manipulation~\cite{kalashnikov2018qt,jaritz2018end,fei2020learn}, robot navigation~\cite{kulhanek2019vision,kretzschmar2016socially}, robot competition~\cite{abreu2019learning} and other robotic control tasks~\cite{lin2020exploration,khan2020graph,luo2019reinforcement,nagabandi2018neural}). However, as a matter of fact, the RL applications in the context of robotics suffer from the poor sample efficiency of RL~\cite{kaiser2019model}. For instance, even when solving a simple task, RL still needs substantial interaction data for policy improvement. Furthermore, the poor sample efficiency not only slows down the policy improvement but also brings about other deleterious problems for deep RL, such as memorization and sensitivity to out-of-manifold samples~\cite{mcallister2019robustness,loquercio2020general}. Generally, RL agent gathers data for policy improvement along the learning process, which means that at the early stage of learning, the amount of training data is small, and deep neural network is prone to memorize~(instead of generalizing from) these training data~\cite{zhang2016understanding}. Unfortunately, such memorization causes the bootstrapping value function does not generalize well to the unvisited~(out-of-manifold) state-action combinations and thus can hamper the policy improvement via Bellman backup~\cite{sutton2018reinforcement}. Moreover, it also degrades the performance of the agent in the environment, due to its exploring preference towards unvisited states.







\begin{figure}
    \centering
    \includegraphics[clip, trim=0 360 550 0, width=0.8\linewidth]{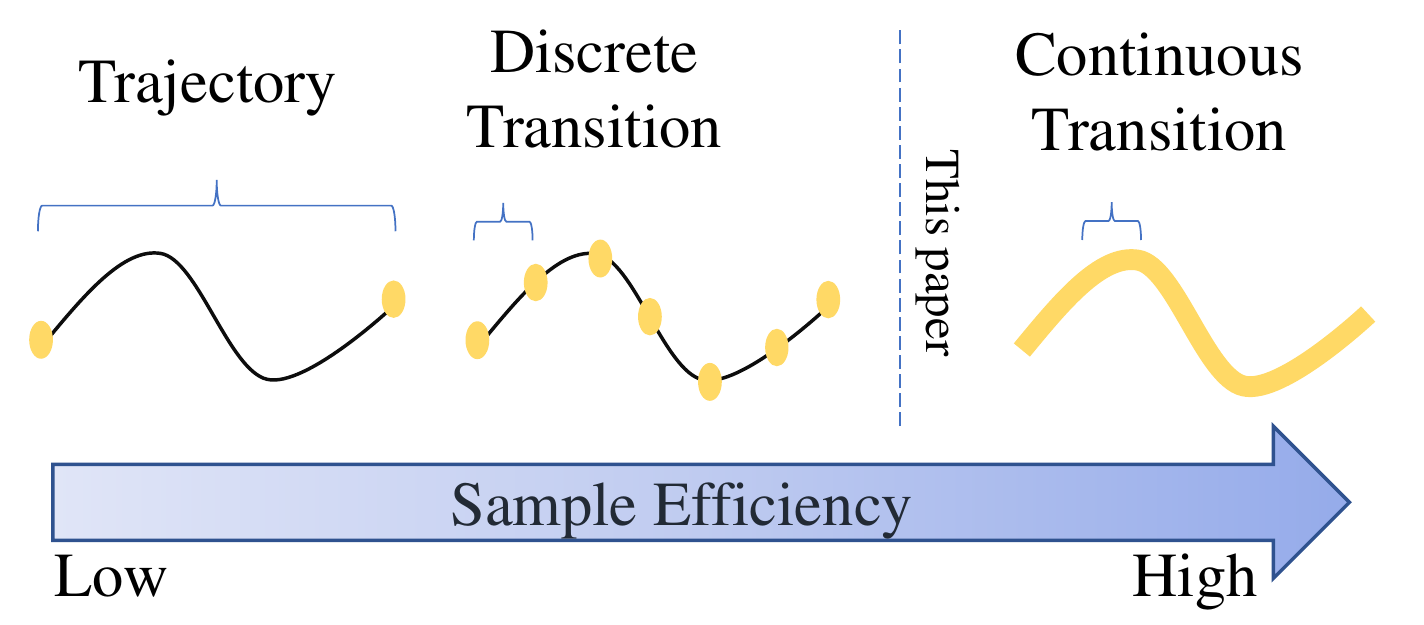}
    \vspace{-5pt}
    \caption{Improve sample efficiency from trajectory to \textit{continuous transition}. From left to right, there are the trajectory, the discrete transition, and the \textit{continuous transition} by interpolating the \ijf{consecutive} discrete transitions.}
    \label{fig:motivation}
    \vspace{-20pt}
\end{figure}

There are many attempts in RL to improve the sample efficiency. One of the most successful attempts is to adopt off-policy instead of on-policy reinforcement learning. On-policy methods~(e.g. PPO~\cite{schulman2017proximal}, TRPO~\cite{schulman2015trust}, A3C~\cite{mnih2016asynchronous}) requires new trajectories to be collected for each update to the policy\ijf{, where a trajectory is comprised of a sequence of states, actions, rewards and done signals, i.e. $(s_0, a_0, r_0, d_0, s_1, ..., d_T)$}. Off-policy methods, such as DDPG~\cite{lillicrap2015continuous}, TD3~\cite{fujimoto2018addressing}, and SAC~\cite{haarnoja2018soft,haarnoja2018soft2} achieve greater sample efficiency via reusing collected data. In these methods, the imprint of the policy is erased from the data by splitting the trajectory data into a set of discrete transitions in terms of action \ijf{(i.e., $\{(s_t, a_t, s_{t+1}, d_t, r_t)\}_{t=0:T-1}$)} so that the transitions are only dependent on the immediate states and actions and can be reused~\cite{sutton2018reinforcement}. However, splitting an episode into several discrete transitions still has a limited improvement on sample efficiency, because the number of transitions is usually small, and only several joint states \ijf{(i.e., $\{s_t\}_{t=0:T}$)} are visible to the policy during training, as depicted in Fig.~\ref{fig:motivation}.






Seeing the great success in improving sample efficiency through decomposing the trajectory into discrete transitions, we wonder if the extracted transitions can be denser and more states of the trajectory can be used to improve the policy. Conventionally, the most popular strategy to extend data is through data-augmentation. Different data-augmentation strategies have been broadly studied and proved to be effective in the field of computer vision~\cite{simard2003best,cirecsan2011high,ciregan2012multi,tasci2015imagenet,krizhevsky2012imagenet}. A recently proposed such technique, MixUp~\cite{zhang2017mixup}, is a simple yet very effective, data augmentation approach for enhancing the performance of deep classification models. Through linearly interpolating random data sample pairs and their training targets, MixUp generates a synthetic set of training samples and use these out-of-manifold examples to regularize the deep models. These methods reveal the potential of improving the performance of deep neural networks by constructing data from those already exists. However, most of these strategies are developed for image inputs~\cite{kostrikov2020image}, and there are still no effective approaches to augment the feature input in RL, to our best knowledge.




In this paper, we exploit the potential of improving sample efficiency in RL by further extracting transitions from the trajectory data via MixUp-alike strategy -- linearly interpolating two \ijf{consecutive transitions. This strategy is aimed at producing meaningful transitions on continuous robotic control tasks. As on the continuous robotic control tasks, the successive states, actions, and rewards are usually similar and could perform transitions smoothly.} In this sense, the constructed transitions by linear interpolating two \ijf{consecutive} transitions might exist in or not far away from the actual transitions manifolds. By this means, we can extract an infinite amount of transitions from the trajectory which varies continuously. 
We name the constructed transition as ``\textit{continuous transition}" compared to the original discrete transition, as depicted in Fig.~\ref{fig:motivation}. 

In our continuous transition framework, an interpolation ratio is used to weigh the importance of the two \ijf{consecutive} transitions. As the interpolation ratio close to 0 or 1, \textit{continuous transitions} are more likely to exist in the authentic dynamics manifolds. The original discrete transitions can be obtained by taking the interpolation ratios as either 0 or 1. The interpolation ratio is sampled from a beta distribution $\mathbb{B}(\beta, \beta)$ parameterized by the temperature $\beta$. When the temperature is close to 1, the beta distribution is similar to uniform distribution $U[0,1]$, and when it is close to 0, the beta distribution is similar to a 2-point Bernoulli distribution. Therefore, tuning the temperature can change the distribution of the interpolation ratio, and thus will change the expectation distance between the \textit{continuous transitions} and the authentic transition manifold. To keep the distance small, we develop an energy-based discriminator to auto-tune the temperature. 
Specifically, the discriminator learns a dynamics model using the authentic discrete transitions. Therefore, the discriminator tends to render a small estimation error if the transition shares the same dynamics as the authentic transition. Therefore, the prediction error between the estimated and the actual dynamics of the transition is used to represent the distance between that transition and the authentic transition manifold. This discriminator guides the temperature $\beta$ to ascend to increase the variety of transitions if the \textit{continuous transitions} are not far away from the estimated transition manifold in expectation. 

The \textbf{main contribution} is summarized as two-fold: i) We propose to construct \textit{continuous transition} to further improve the sample efficiency of the reinforcement learning on continuous control problems; ii) To keep the \textit{continuous transition} not far from the authentic transition manifold, we introduce an energy-based discriminator to auto-tune the temperature of the beta distribution. Extensive and comprehensive experimental evaluations demonstrate the our proposed framework achieves clear improvement to sample efficiency on complex continuous control tasks.


\section{RELATED WORK}

Improving sample efficiency is one of the most critical issues in reinforcement learning, which has been widely studied. Several aspects could alleviate the limitations: considering the exploration-learning procedure, how to better optimize the policy~\cite{schulman2015trust,florensa2017stochastic,wang2018noisy} and how to efficiently explore the environment~\cite{plappert2017parameter,singh2010intrinsically,pathak2017curiosity}; considering the environment, how to learn the dynamics model~\cite{tamar2016value,racaniere2017imagination,pong2018temporal}, etc.

\textbf{Improving Sample Efficiency.} One cause for the poor sample efficiency of deep RL methods is on-policy learning: some of the most commonly used deep RL algorithms, such as TRPO~\cite{schulman2015trust}, PPO~\cite{schulman2017proximal} or A3C~\cite{mnih2016asynchronous}, require new samples to be collected for each gradient step. This quickly becomes extremely expensive, as the number of gradient steps and samples per step needed to learn an effective policy increases with task complexity. Off-policy algorithms aim at reusing past experience, e.g., DQN~\cite{mnih2015human}. Extending these methods to continuous control, deep deterministic policy gradient (DDPG)~\cite{lillicrap2015continuous}, provides for sample-efficient learning but is notoriously challenging to use due to the brittleness of learning and hyperparameter sensitivity~\cite{duan2016benchmarking}. ~\cite{fujimoto2018addressing} combines twin critics to further improve its stability. To address the efficiency of exploration, ~\cite{haarnoja2018soft,haarnoja2018soft2} resort to maximizing both value and policy entropy.

\textbf{Data Augmentation.} Data augmentation plays an important role on almost all successful applications of deep learning, ranging from image classification~\cite{simard2003best,cirecsan2011high,ciregan2012multi,tasci2015imagenet,krizhevsky2012imagenet} to speech recognition~\cite{graves2013speech,amodei2016deep}. In these works, substantial domain knowledge is leveraged to design suitable data transformations leading to an improved generalization. In image classification, for example, rotation, translation, cropping, resizing and flipping~\cite{lecun1998gradient,simonyan2014very} are routinely used to endow the model with visually plausible invariances through the training data. As for speech recognition, it is prevalent to inject noise to improve the robustness and accuracy of the trained models~\cite{amodei2016deep}. Recently, MixUp~\cite{zhang2017mixup}, a simple yet effective data-augmentation strategy, synthesize out-of-manifold training data via linear interpolating random data sample pairs and their targets. In ~\cite{zhang2017mixup}, MixUp is shown to efficiently improve the classification accuracy of the deep neural networks. To remedy the out-of-manifold problem, that is, the synthesized data conflict too much with the authentic training data, ~\cite{guo2019mixup} proposed to classify the synthetic data and original data. However, as noted in ~\cite{kostrikov2020image}, these tasks are invariant to certain image transformations such as rotations or flips. In the field of RL, data augmentation is still applied in tasks with image input~\cite{cobbe2019quantifying,kostrikov2020image,laskin2020reinforcement,srinivas2020curl,raileanu2020automatic}. Data augmentation on feature input in RL is still touchless, to our best knowledge.


\section{PRELIMINARIES}


\textbf{Reinforcement Learning.} We consider a standard RL framework where an agent interacts with an environment according to the observation in discrete time. Formally, the agent observes a state feature $s_t$~(e.g., position, velocity) from the environment at each time step $t$ and then chooses an action $a_t$ according to its policy $\pi$. The environment returns a reward $r_t$ and the next state $s_{t+1} \sim p_e(s_t, a_t)$, where $p_e$ is the environment transition mapping. Normally, the environment will also provide a done signal $d_t$ to indicate the end of the trajectory. Transition $T_t$ is the quintet $(s_t, a_t, r_t, s_{t+1}, d_t)$. Conventionally, a trajectory is \ijf{composed} of successive transitions. And the return of a trajectory from timestep $t$ is $\eta_t=\sum_{k=0}^\infty \gamma^k r_{t+k}$, where \ijf{$\gamma \in [0, 1)$} is the discount factor. RL aims to optimize the policy to maximize the expected return from each state $s_t$.

\textbf{MixUp.} MixUp~\cite{zhang2017mixup}, a data-augmentation strategy, was first introduced to image classification. In a nutshell, MixUp enhances the training of deep classification models by generating synthetic samples through linearly interpolating a pair of training samples as well as their modeling targets. Given a pair of samples $x^i$ and $x^j$ from the original training set, where $x$ includes the input and the one-hot encoding of the corresponding class of the sample. The linear interpolation w.r.t. an interpolation ratio $\epsilon \in [0, 1]$ is formulated as:
\begin{equation}
    \mathbb{M}_\epsilon(x^i, x^j) = \epsilon x^i + (1-\epsilon) x^j,
\end{equation}
where the interpolation ratio $\epsilon$ weighs the importance of the samples. The synthetic data are then fed into the model for training to minimize the loss function. MixUp employs these out-of-manifold samples to regularizes the deep models.

\section{METHODOLOGIES}

Generally, there is a minimal time-interval for an agent to make two successive decisions and a trajectory is composing of several successive discrete transitions with the minimal time-interval. These transitions are decoupled from the behavior policy and thus they can be repeatedly used to improve the policy~\cite{mnih2015human}. Off-policy methods like DDPG~\cite{fujimoto2018addressing} and SAC~\cite{haarnoja2018soft,haarnoja2018soft2} training with transitions largely improve sample efficiency on complex continuous control tasks, against the on-policy methods~\cite{schulman2017proximal,schulman2015trust,mnih2016asynchronous} which are training with trajectory data.

\begin{algorithm}[t]  
    \caption{Continuous Transitions Framework}\label{alg:ctf}   
    \textbf{Input:}initial beta distribution parameters $\beta$, $Q$-function parameters $\theta$, replay buffer $\mathcal{B}$, tolerance $m$.
    
    \textbf{Output:} $Q$-function $Q_\theta$.
        
    \begin{algorithmic}[1]

     \For{each iteration}
        \State Sample a batch of pairs of \ijf{consecutive} transitions $(T_t, T_{t+1})$ from the replay buffer $\mathcal{B}$.
        \State Sample a batch of interpolation ratio $\{\epsilon \sim \mathbb{B(\beta, \beta)}\}$
        \begin{varwidth}[t]{\linewidth}
        \State Construct \textit{continuous transitions}: \par
        \hskip\algorithmicindent\hskip\algorithmicindent $\{T'_t \gets \mathbb{M}_\epsilon(T_t,T_{t+1})\}$ 
        \end{varwidth}
        
        \State Update $Q_{\theta}$ with {$T'_t$} \par
        \algorithmiccomment{\ijf{Any Method using Bellman Backup}} 
    
        \State Update $\phi$ by one gradient descent step on Equ.~(\ref{equ:disc})
        \State Update $\beta$ by one gradient descent step on Equ.~(\ref{equ:att})
    \EndFor
\State\Return $Q_\theta$
    
    \end{algorithmic} 
\end{algorithm}

\begin{figure}
    \centering
    \includegraphics[clip, trim=0 300 320 0, width=0.8\linewidth]{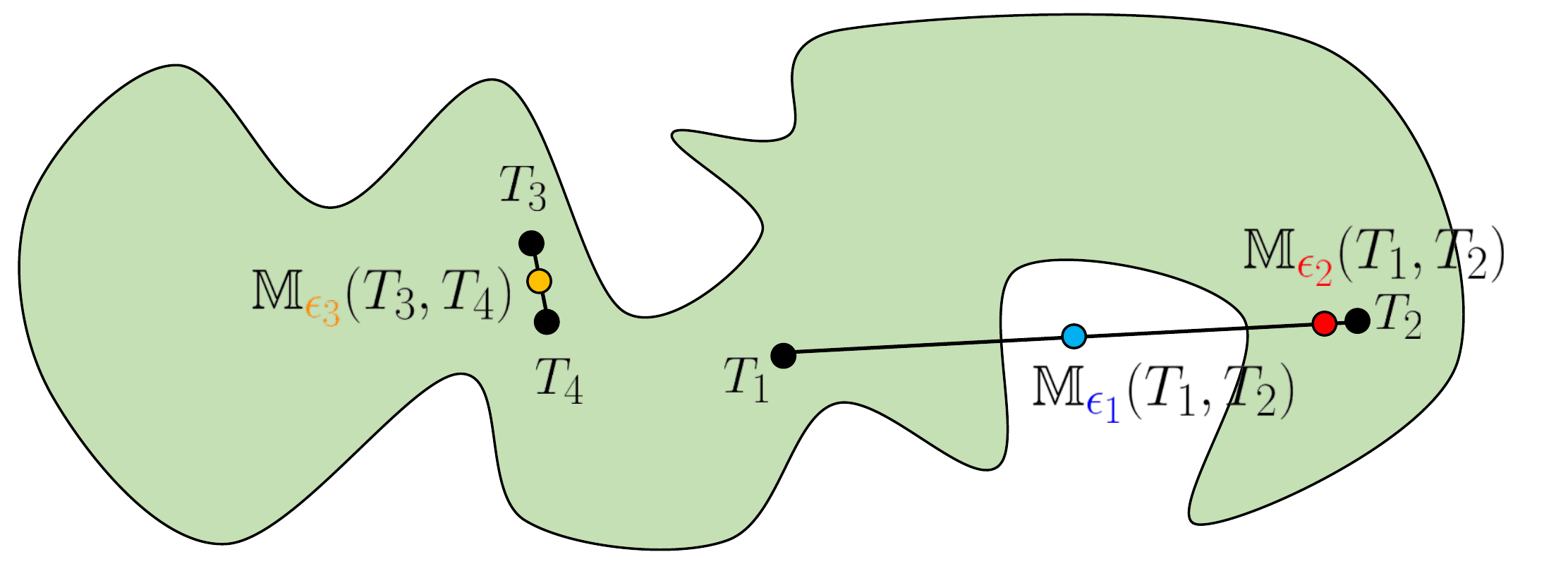}
    \vspace{-10pt}
    \caption{The sketch of the transition manifold. Constructed transition by linearly interpolating close transitions~(e.g., the orange dot in the figure) is likely to be in-manifold while interpolating distant transitions~(e.g., the blue dot) is likely to be out-of-manifold. When the interpolation ratio is close to 0 or 1, the constructed transition is similar to the original discrete transition~(e.g., the red dot in the figure).}
    \label{fig:transition}
    \vspace{-15pt}
\end{figure}

 \textbf{Continuous Transition}. However, improving policy with transitions still has its limitation in enhancing the sample efficiency, because the amount of transitions is usually small and the value assignment only happens to their joint states. The intermediate information between these joint states is uncollected and largely wasted. Obtaining these intermediate information might help the policy to better generalize instead of memorizing the information of the several discrete joint states. Moreover, it might also help the value to propagate across different trajectories if these trajectories intersect at similar intermediate states. Therefore, in this paper, we aims to exploit the potential of the intermediate information along the trajectory to further improve the sample efficiency.

To obtain the intermediate information for training, we resort to linear interpolating two \ijf{consecutive} discrete transitions to construct new transitions that varies smoothly and continuously along the trajectory, as depicted in Fig.~\ref{fig:motivation}. This synthesized transition is termed as \textit{continuous transition}. Specifically, a \textit{continuous transition} $T^\prime_t$ is obtained by $T^\prime_t = \mathbb{M}_\epsilon(T_t, T_{t+1})$, where $\epsilon$ is the interpolation ratio sampled from a beta distribution, i.e., $\epsilon \sim \mathbb{B}(\beta, \beta)$ (parameterised by temperature $\beta \in (0, 1]$). The reason why the transitions to interpolate should be \ijf{consecutive} is that the \ijf{consecutive} transitions usually has more similar dynamics information (e.g., position) on continuous control tasks. As depicted in Fig.~\ref{fig:transition}, the interpolation of two close transitions is more likely to produce a new transition~(yellow dot in the figure) which \ijf{also exists} in the manifold, compared to interpolation between two distant transitions~(blue dot in the figure). Then, the \textit{continuous transitions} instead of the original discrete transitions are used for training.


\begin{figure*}[t]
\centering
\begin{subfigure}{.245\textwidth}
  \centering
  \includegraphics[width=\linewidth]{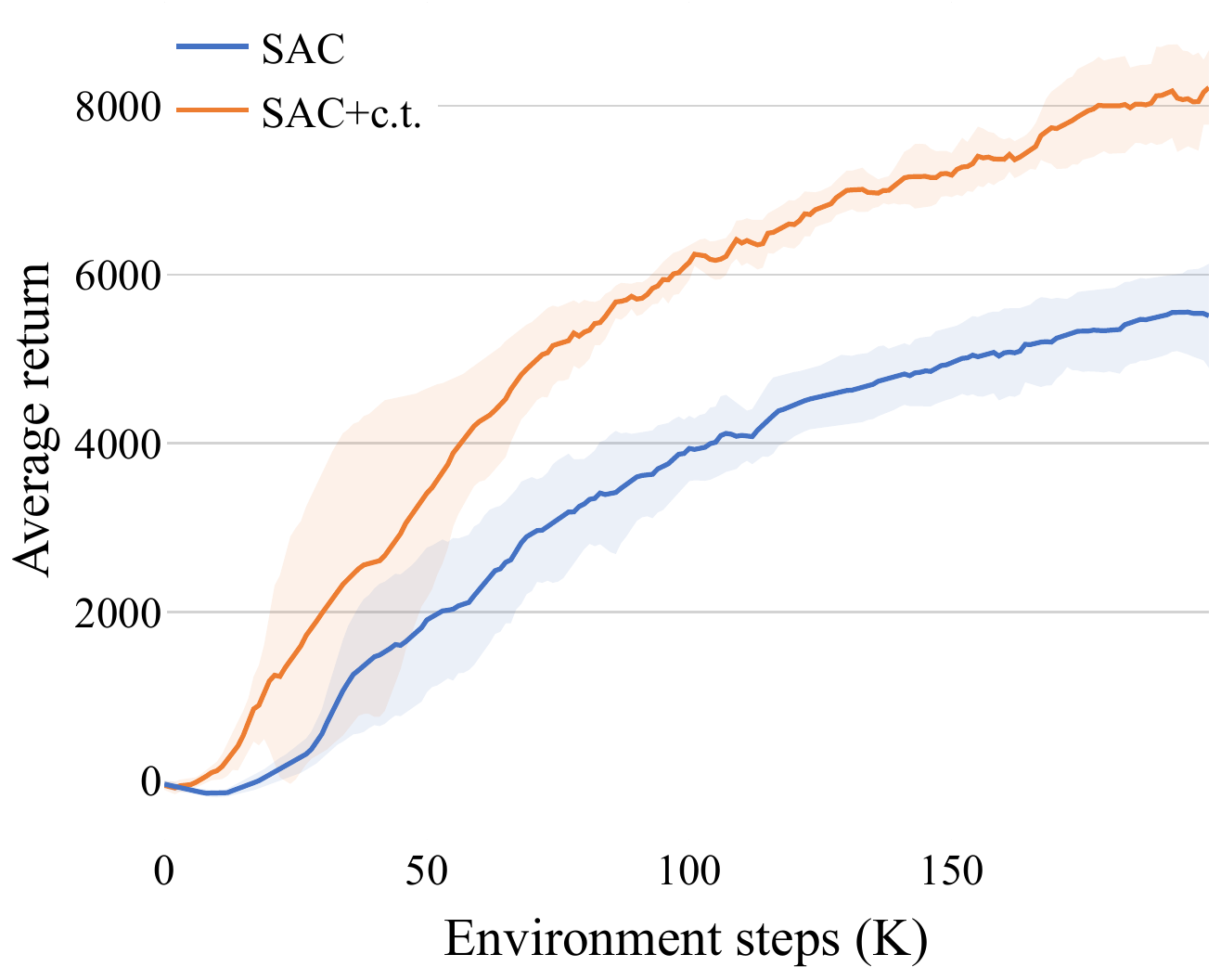}
  \caption{Cheetah}
  \label{fig:sac_cheetah}
\end{subfigure}
\begin{subfigure}{.245\textwidth}
  \centering
  \includegraphics[width=\linewidth]{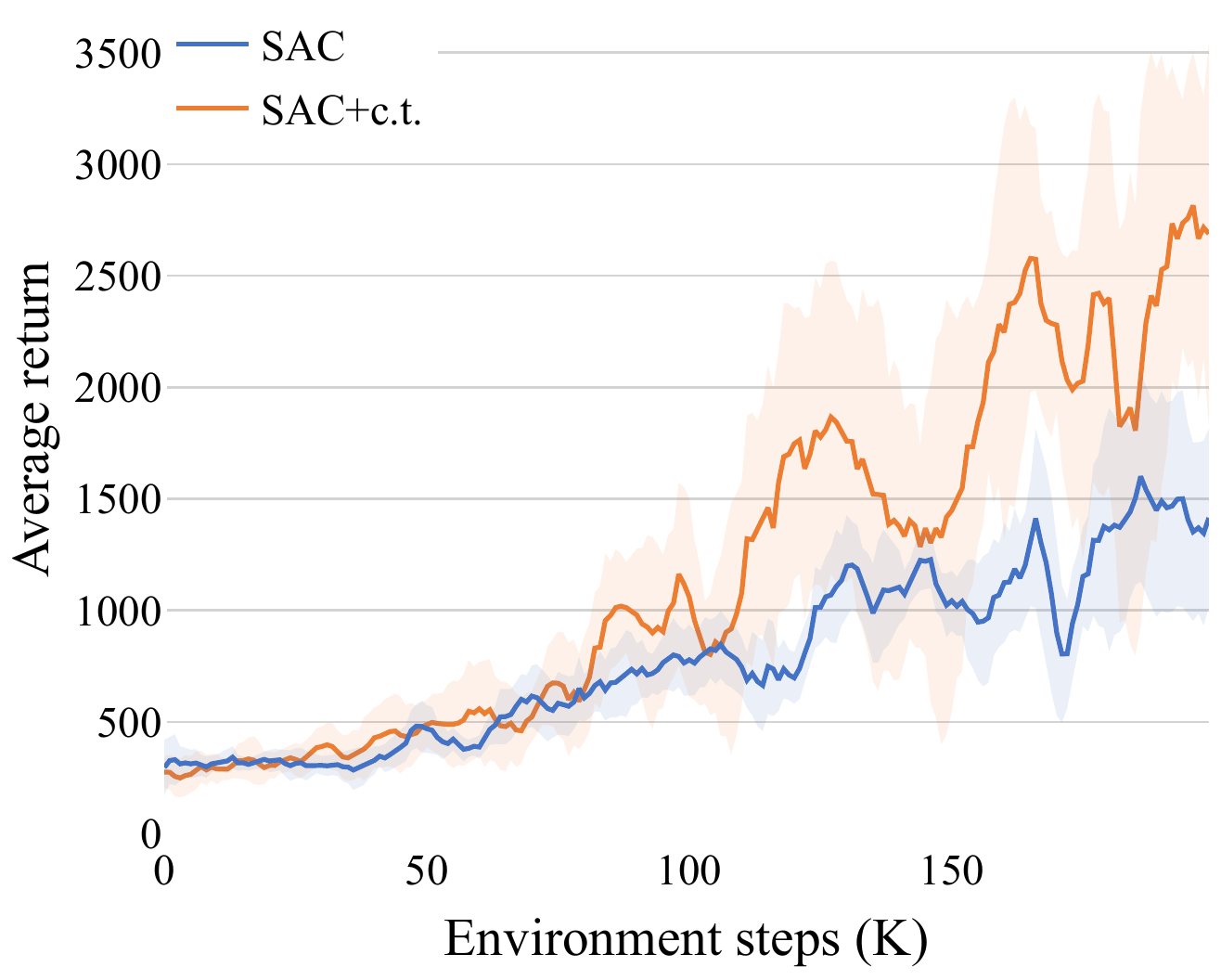}
  \caption{Walker}
  \label{fig:sac_walker}
\end{subfigure}
\begin{subfigure}{.245\textwidth}
  \centering
  \includegraphics[width=\linewidth]{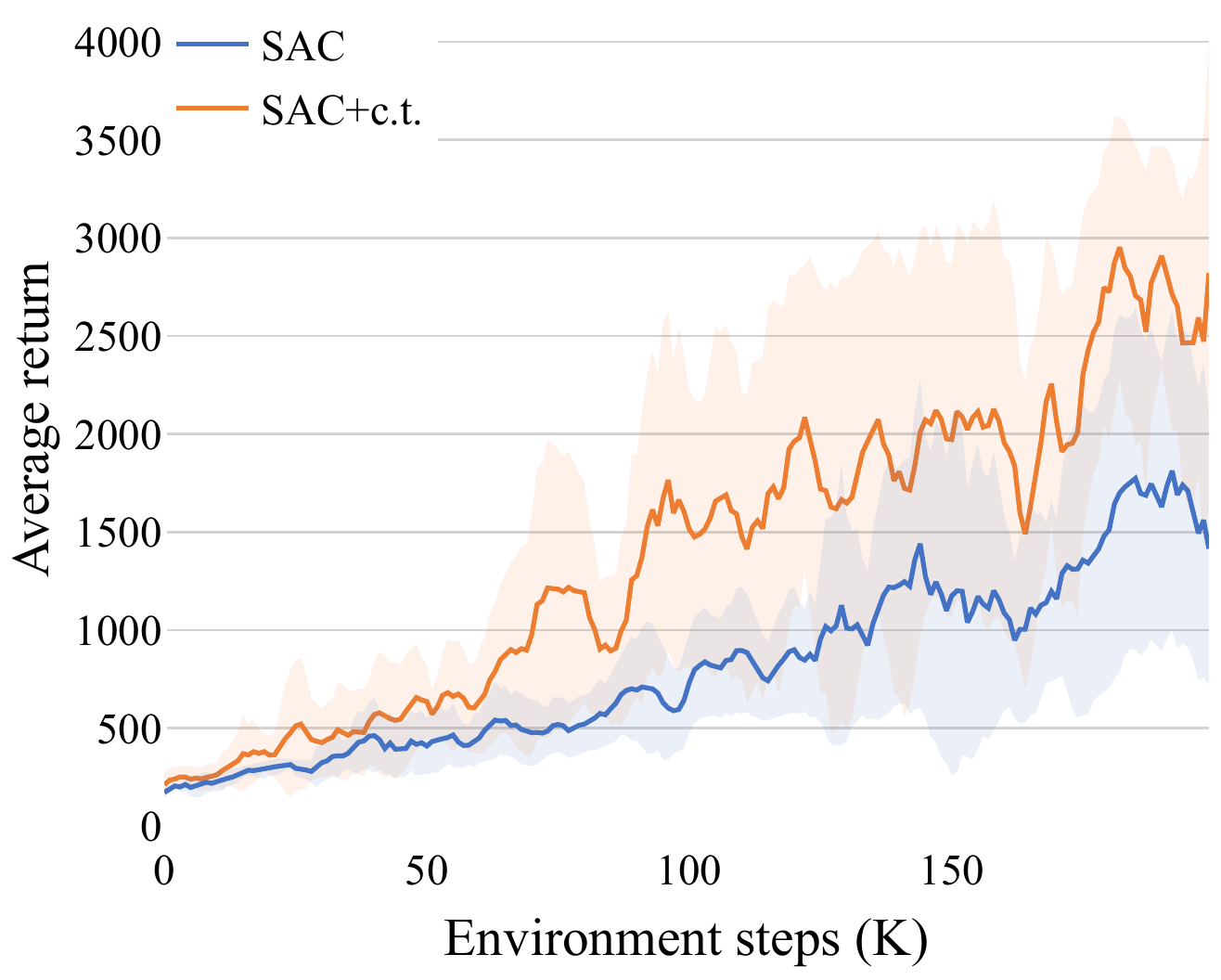}
  \caption{Hopper}
  \label{fig:sac_hopper}
\end{subfigure}
\begin{subfigure}{.245\textwidth}
  \centering
  \includegraphics[width=\linewidth]{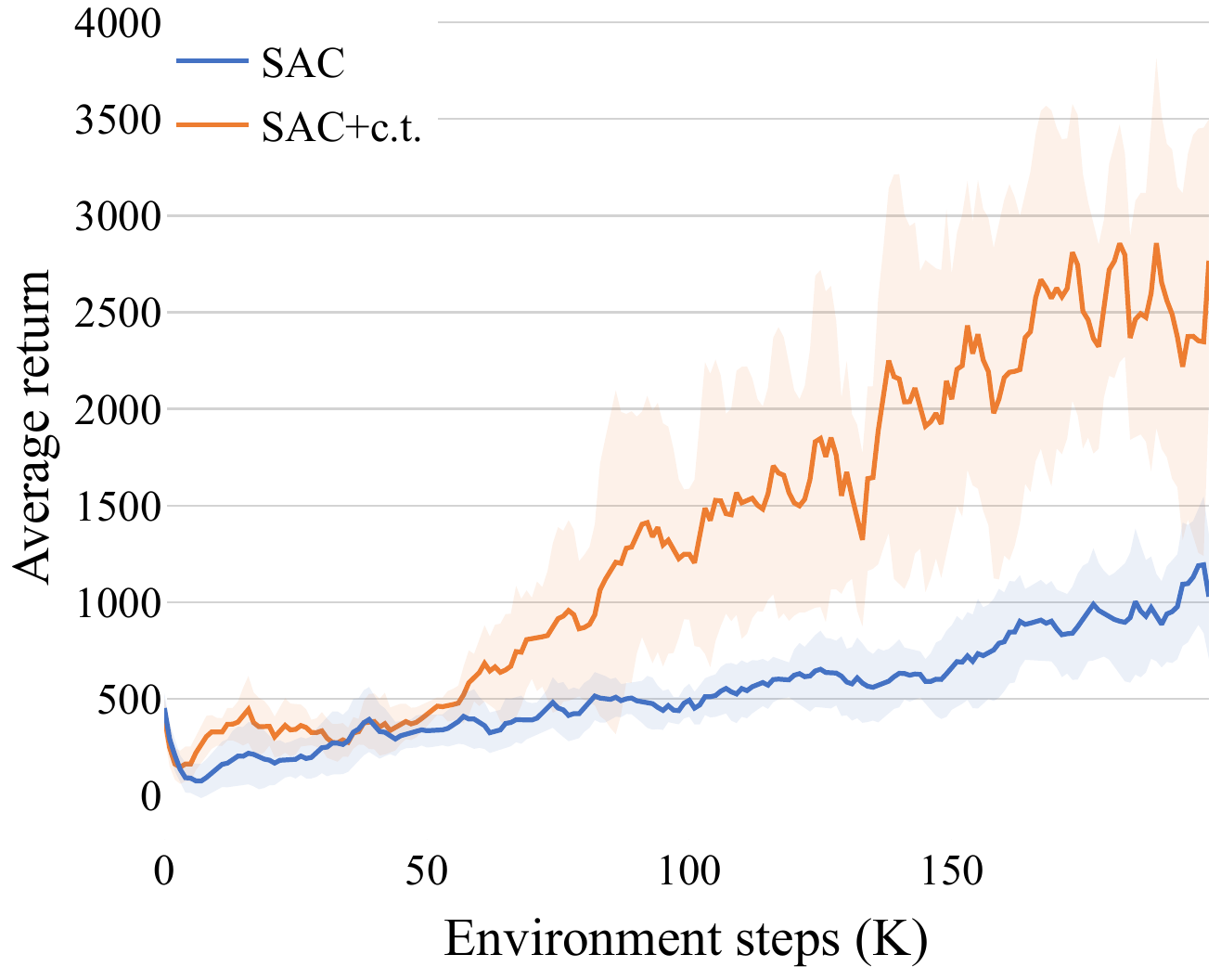}
  \caption{Ant}
  \label{fig:sac_ant}
\end{subfigure}%
\vspace{-5pt}
\caption{Evaluation curves of SAC~\cite{haarnoja2018soft2} and SAC+c.t., where SAC+c.t. consistently improves the performance of SAC.}
\vspace{-10pt}
\label{fig:sac}
\end{figure*}

\begin{figure*}[t]
\centering
\begin{subfigure}{.245\textwidth}
  \centering
  \includegraphics[width=\linewidth]{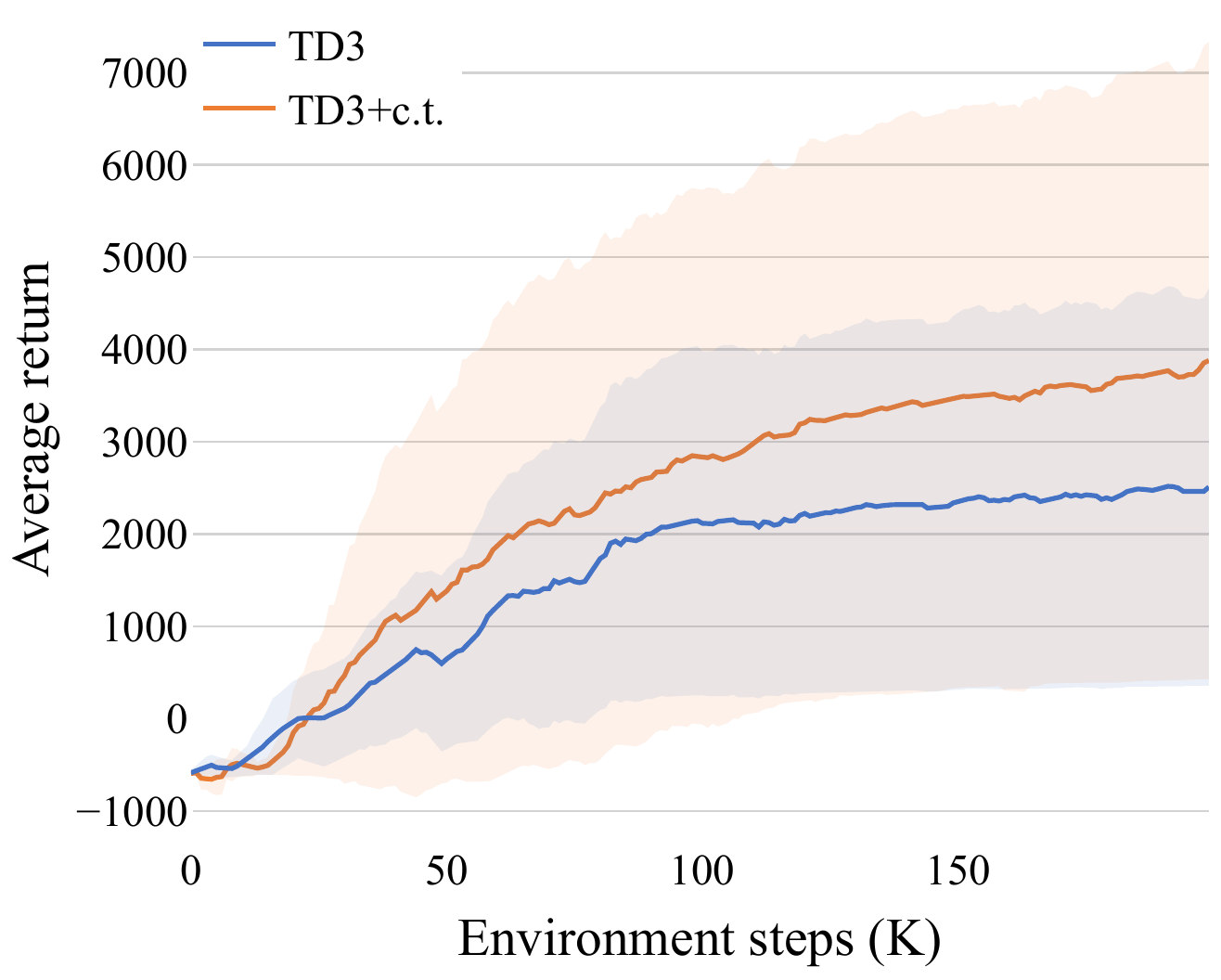}
  \caption{Cheetah}
  \label{fig:td3_cheetah}
\end{subfigure}
\begin{subfigure}{.245\textwidth}
  \centering
  \includegraphics[width=\linewidth]{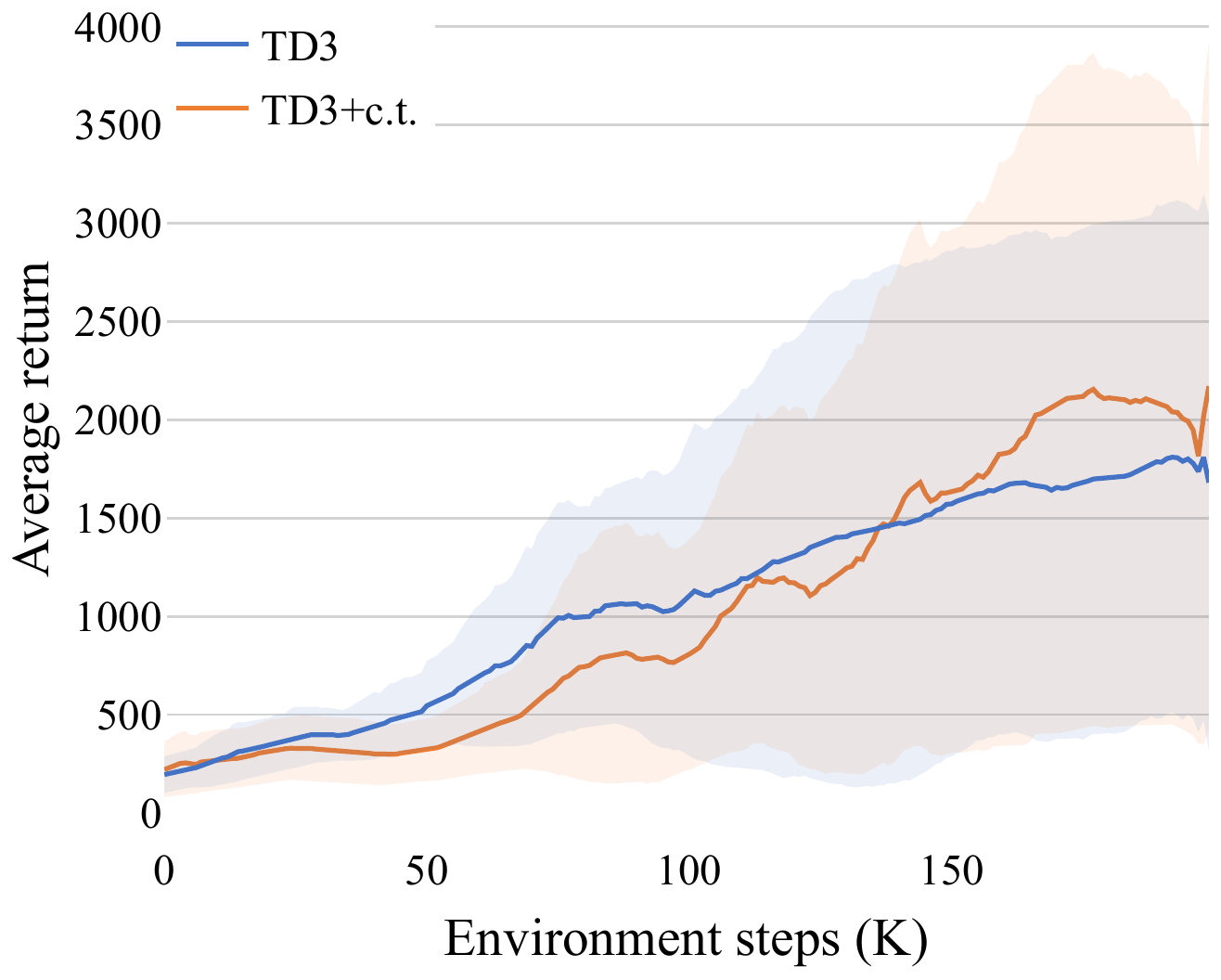}
  \caption{Walker}
  \label{fig:td3_walker}
\end{subfigure}
\begin{subfigure}{.245\textwidth}
  \centering
  \includegraphics[width=\linewidth]{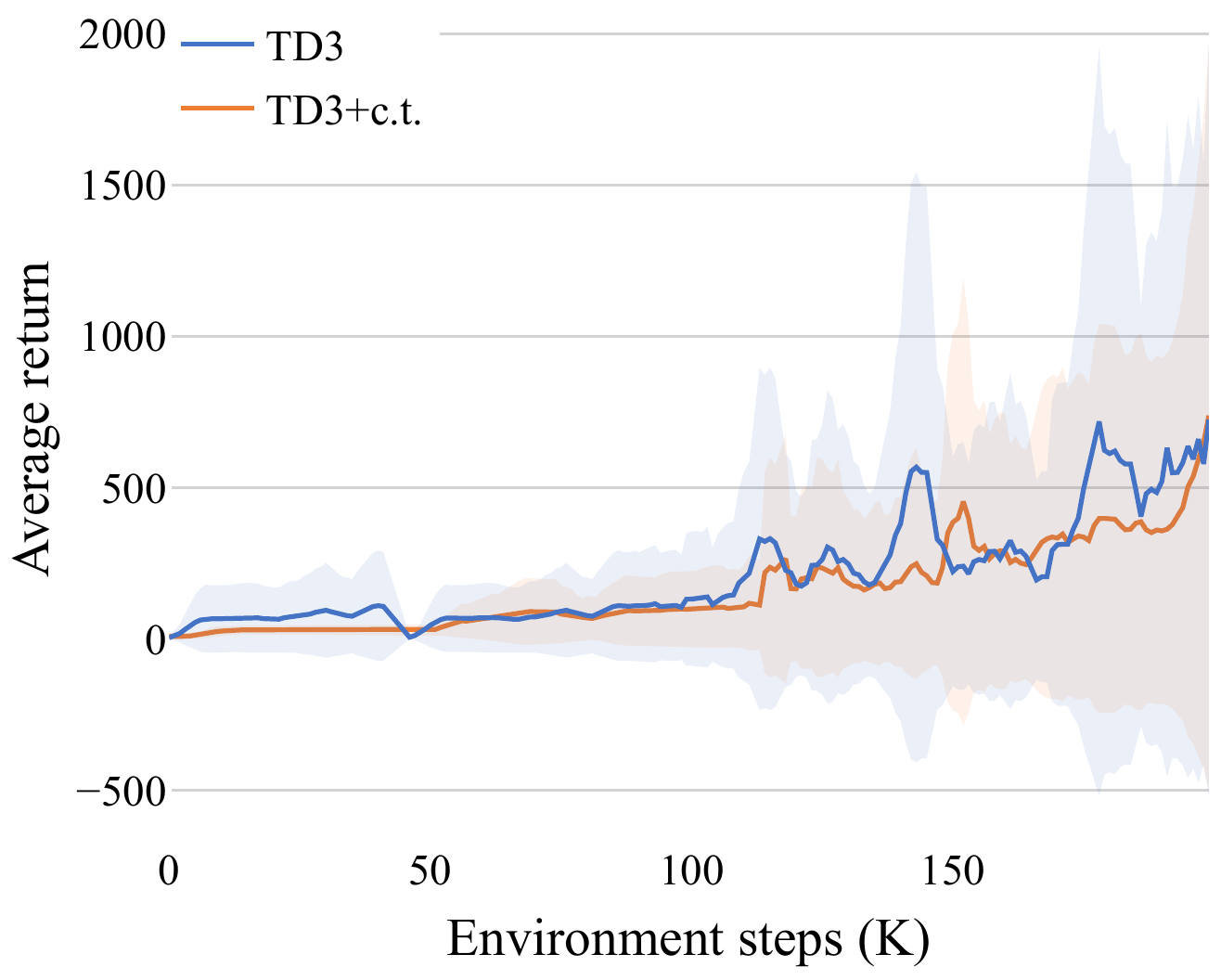}
  \caption{Hopper}
  \label{fig:td3_hopper}
\end{subfigure}
\begin{subfigure}{.245\textwidth}
  \centering
  \includegraphics[width=\linewidth]{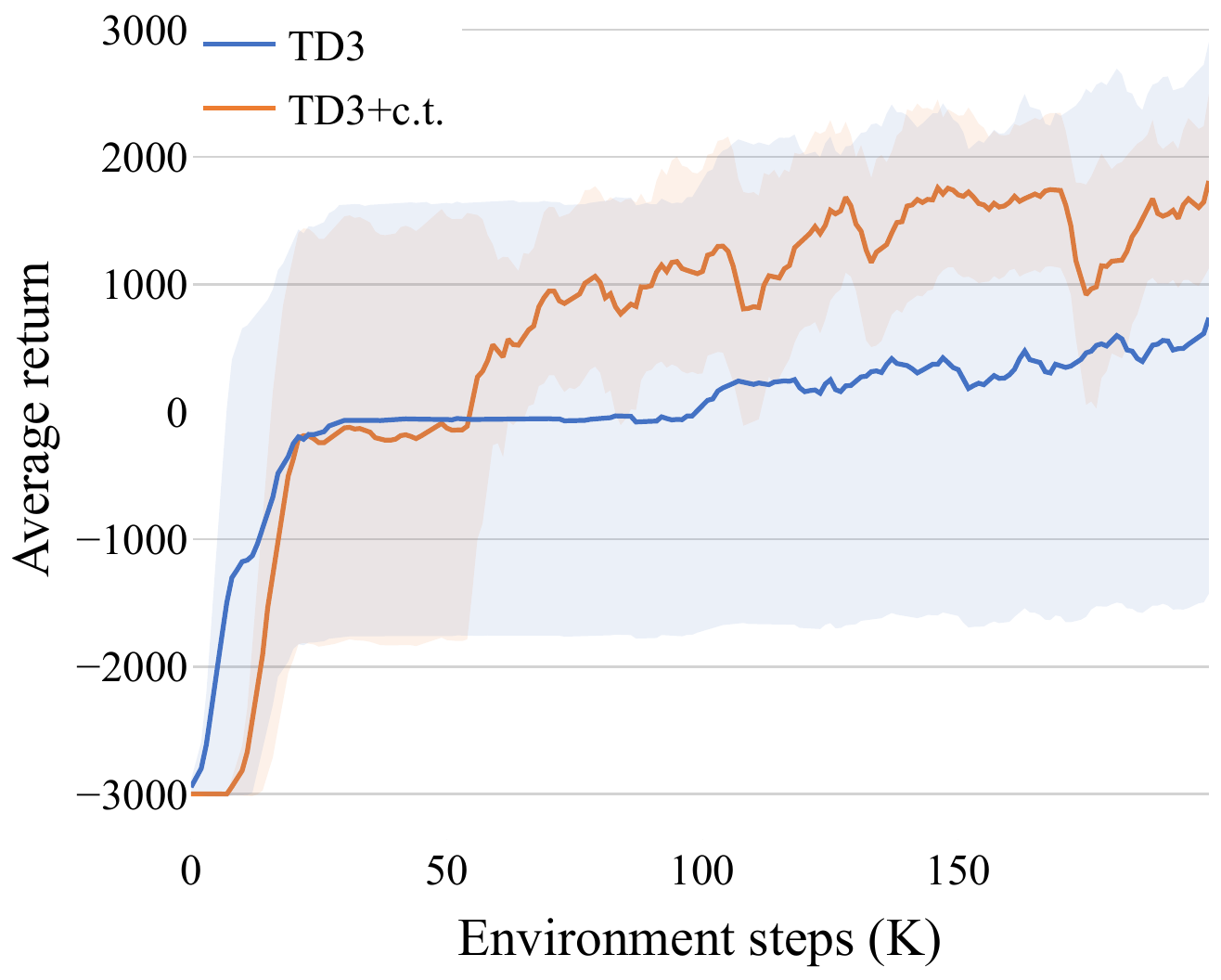}
  \caption{Ant}
  \label{fig:td3_ant}
\end{subfigure}%
\vspace{-5pt}
\caption{Evaluation curves of TD3~\cite{fujimoto2018addressing} and TD3+c.t., where TD3+c.t. consistently improves the performance of TD3.}
\vspace{-15pt}
\label{fig:td3}
\end{figure*}

\textbf{Automatic Temperature Tuning}. In the previous section, we derived a linear interpolation algorithm for constructing \textit{continuous transitions}. The most important hyperparameter is the temperature $\beta \in (0, 1]$ of the beta distribution. When $\beta$ is close to 0, the beta distribution is similar to a 2-point Bernoulli distribution. In this scenario, the interpolation ratio sampled from the beta distribution is about 0 or 1, reducing the \textit{continuous transition} to one of the discrete transitions~(red dot in Fig.~\ref{fig:transition}). When $\beta$ is approaching 1, the beta distribution is similar to a uniform [0, 1] distribution. In this case, the \textit{continuous transition} is likely \ijf{to} be out-of-manifold~(blue dot in Fig.~\ref{fig:transition}). In this sense, the expectation distance between the \textit{continuous transitions} and the authentic transitions manifold is positively \ijf{associated} to the value of $\beta$. We could adjust the $\beta$ to keep the expectation distance below the tolerance $m$.

As for distance modeling, we adopt an energy-based discriminator~\cite{zhao2016energy} to estimate the distance between a transition and the transition manifold. We denote $(s_t, a_t)$ as $x_t$ and $(s_{t+1}, r_t, d_t)$ as $y_t$. Then, the distance estimator $d(\cdot)$ is formulated as:
\begin{equation}
\label{equ:disc}
    d(T_t) = ||E(x_t, y_t)|| = ||f_\phi(x_t) - y_t||^2_2,
\end{equation}
where $E$ is the energy of the pair $(x_t, y_t)$ and $f_\phi$ is the MLP~\cite{pal1992multilayer} parameterized by $\phi$. This discriminator is improved by optimizing $\phi$ to minimize Equ.(\ref{equ:disc}) for all discrete transitions. For simplicity, we denote $E_t$ for $E(x_t, y_t)$. 
\ijf{To ensure the estimated distance of the authentic discrete transition to be zero}, we further correct the discriminator by subtracting the energy of the discrete transition:
\begin{equation}
    \tilde{d}(\mathbb{M}_\epsilon(T_t, T_{t+1})) = \tilde{d}(T^\prime_t) =  || E^\prime_t - \mathbb{M}_\epsilon(E_t, E_{t+1})||^2_2,
    \label{equ:relative_ebm}
\end{equation}
where $E'_t$ is the energy w.r.t. the \textit{continuous transition} $T^\prime_t$. When $\epsilon$ is 0 or 1, $\tilde{d}(\mathbb{M}_\epsilon(T_t, T_{t+1}))$ equals to 0, agnostic to the parameter $\phi$. This means that the unbiased distance estimator automatically adapt itself to work even at the very beginning of the training \ijf{(i.e., $f_\phi$ is not necessary to be optimal)}.

Unfortunately, it is non-trivial to choose the optimal $\beta$ so that the expectation distance is kept within a specific tolerance, and the $\beta$ might vary at different periods of the training process. Instead of requiring the user to set the temperature manually, we can automate this process by formulating a constraint satisfaction problem. 
Our aims is to find a stochastic policy to construct \textit{continuous transitions} that the distance between the synthetic transition and the actual \ijf{manifold} is under a tolerance $m$. Formally, we want to solve the constrained optimization problem
\begin{equation}
\begin{aligned}
    {\rm max\ } \beta, {\rm \ s.t. \ } \mathbb{E}[\tilde{d}(\mathbb{M}_\epsilon(T_t, T_{t+1}))] \leq m, 0 < \beta \leq 1,\\
    {\rm where \ } \epsilon \sim \mathbb{B}(\beta, \beta).
\end{aligned}    
\end{equation}
we can optimize the $\beta$ by function approximators and stochastic gradient descent using the simplified \ijf{objective}:
\begin{equation}
\begin{aligned}
\label{equ:att}
    \underset{\beta}{\rm argmin}{\rm \ log}\beta \big(\mathbb{E}[\tilde{d}(\mathbb{M}_\epsilon(T_t, T_{t+1}))] - m\big), 
    {\rm where \ } \epsilon \sim \mathbb{B}(\beta, \beta).
\end{aligned}    
\end{equation}
After each \ijf{gradient descent step}, we clip $\beta$ to be within $(0, 1]$.

\section{\ijf{SIMULATIONS}}




To demonstrate the effectiveness of our introduced \textit{continuous transitions}, we conduct several comprehensive experimental evaluations to compare with the current advanced model-based / model-free baseline methods on difficult continuous control MuJoCO environments~\cite{todorov2012mujoco}. Moreover, we also ablate different choices of the important components and hyperparameters of \textit{continuous transitions} to validate their contributions to the model performance. 


\begin{table}[t]
\centering
\caption{Performance on OpenAI Gym at 200K timesteps. The results show the mean and standard deviation across four runs. For baselines, we report the best number in SUNRISE~\cite{lee2020sunrise}. The best results are highlighted in bold.}
\label{tab:baselines}
\begin{tabular}{l | cccc} 
\toprule
 Method        & Cheetah       & Walker        & Hopper       & Ant         \\ \midrule
METRPO  & 2284$\pm$900  & -1609$\pm$658 & 1273$\pm$501 & 282$\pm$18     \\
PETS & 2288$\pm$1019 & 283$\pm$502   & 115$\pm$621  & 1166$\pm$227 \\
POPLIN-A & 1563$\pm$1137 & -105$\pm$250  & 203$\pm$963  & 1148$\pm$438 \\
POPLIN-P & 4235$\pm$1133 & 597$\pm$479   & 2055$\pm$614 & 2330$\pm$321 \\
SUNRISE   & 5371$\pm$483  & 1926$\pm$695  & 2602$\pm$307 & 1627$\pm$293 \\ \midrule
TD3~\cite{fujimoto2018addressing}      & 2509$\pm$2152  & 1681$\pm$1365 & 725$\pm$1248 & 738$\pm$2165  \\ 
TD3+c.t.  & 3880$\pm$3452  & 2171$\pm$1751  & 1817$\pm$739 & 1232$\pm$685  \\ 
SAC~\cite{haarnoja2018soft2}      & 5512$\pm$616  & 1415$\pm$401  & 1415$\pm$692 & 1029$\pm$322 \\
SAC+c.t.  &  \textbf{8218$\pm$440}  &  \textbf{2688$\pm$860}  &  \textbf{2821$\pm$1179}  &  \textbf{2767$\pm$731} \\ \bottomrule
\end{tabular}
\vspace{-15pt}
\end{table}


\begin{figure*}[t]
\centering
\begin{subfigure}{.245\textwidth}
  \centering
  \includegraphics[width=\linewidth]{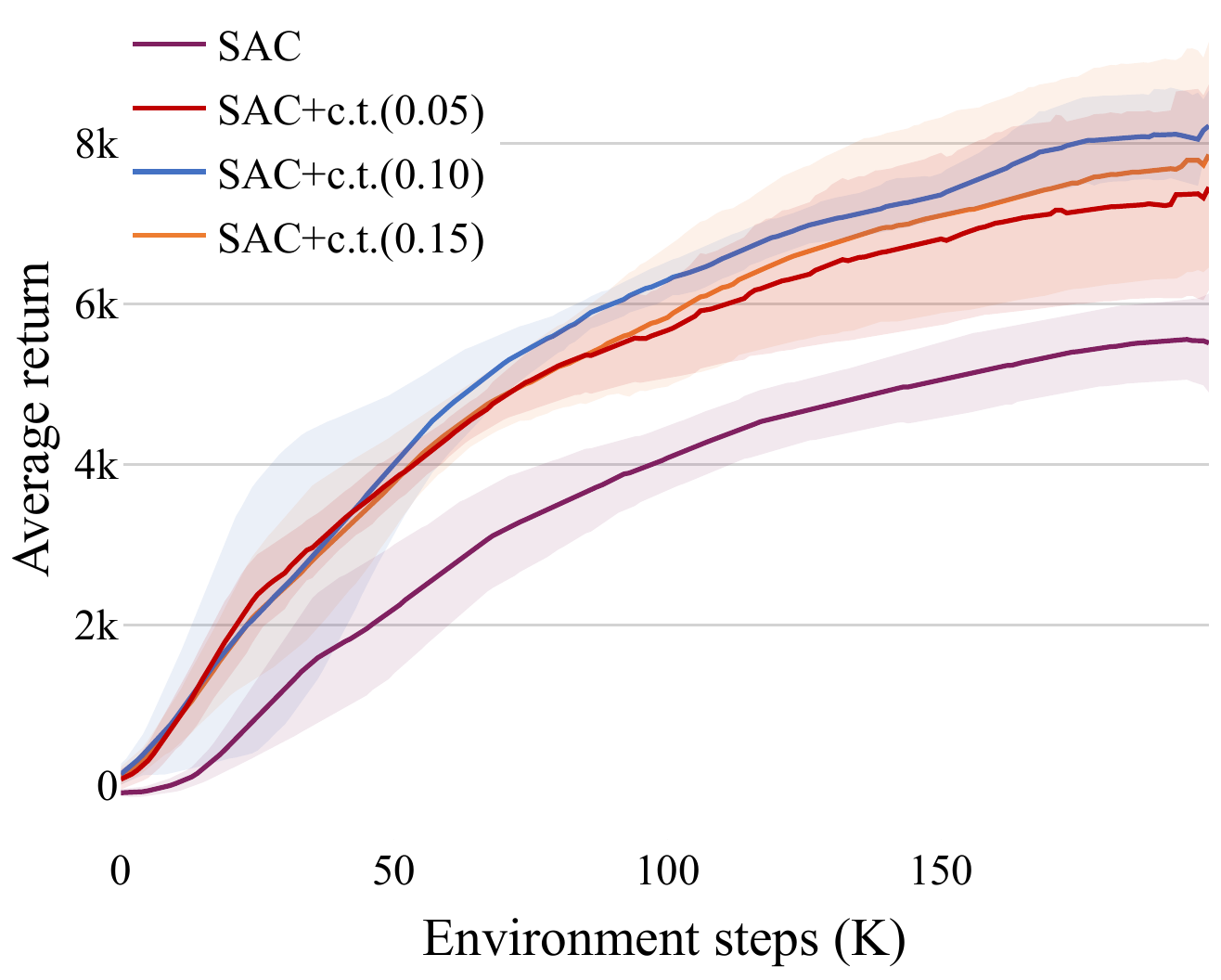}
  \caption{Cheetah}
  \label{fig:tor_cheetah}
\end{subfigure}
\begin{subfigure}{.245\textwidth}
  \centering
  \includegraphics[width=\linewidth]{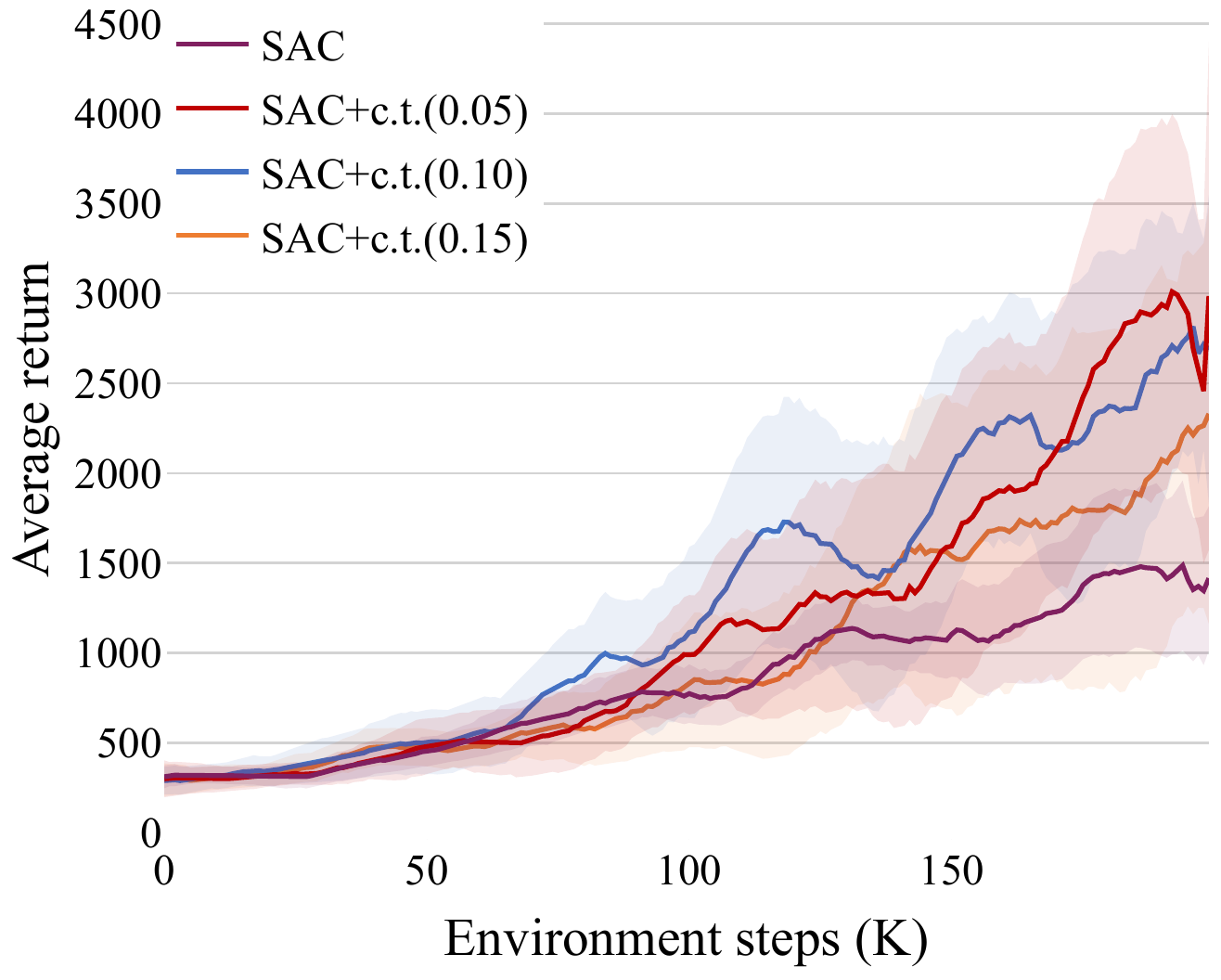}
  \caption{Walker}
  \label{fig:tor_walker}
\end{subfigure}
\begin{subfigure}{.245\textwidth}
  \centering
  \includegraphics[width=\linewidth]{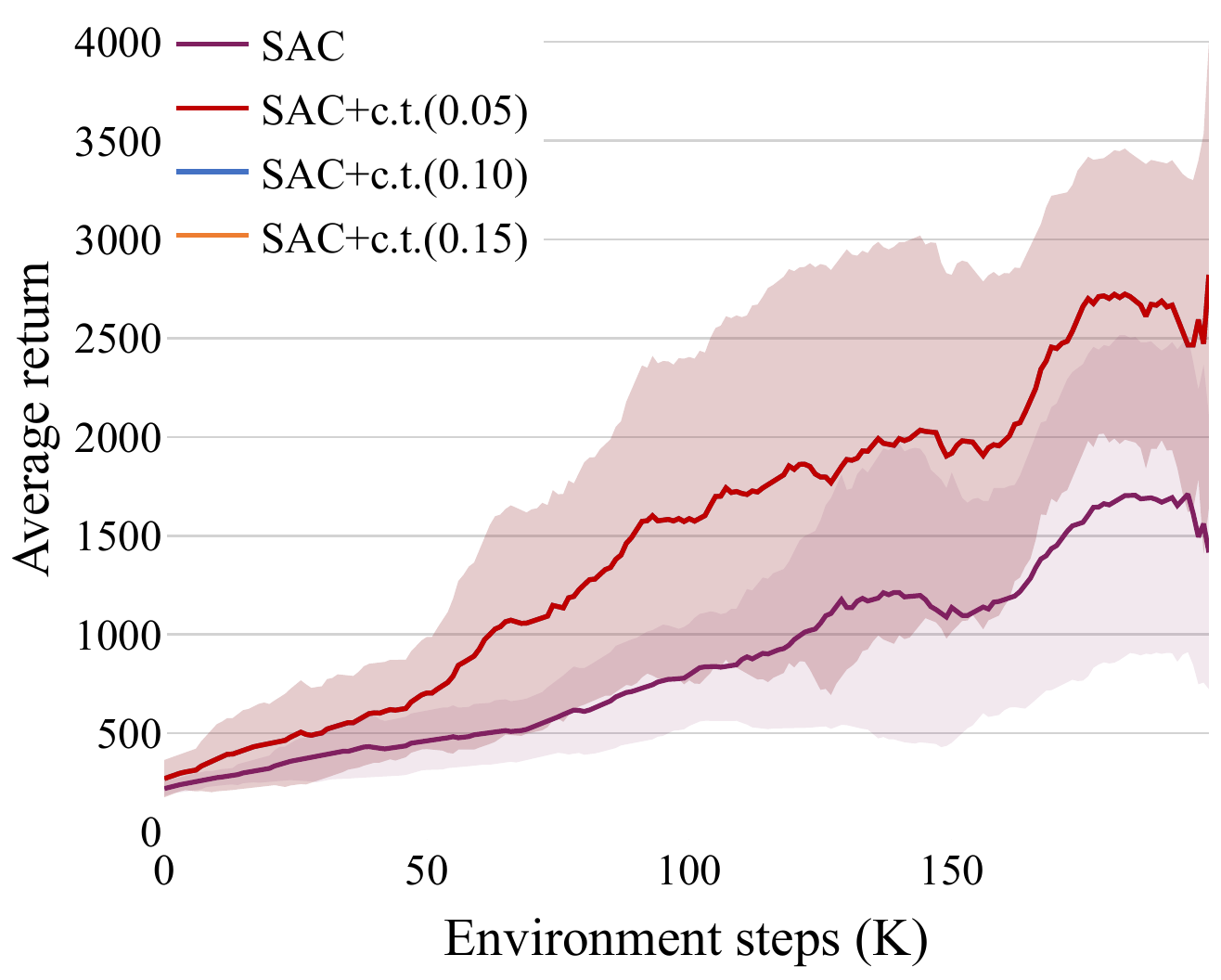}
  \caption{Hopper}
  \label{fig:tor_hopper}
\end{subfigure}
\begin{subfigure}{.245\textwidth}
  \centering
  \includegraphics[width=\linewidth]{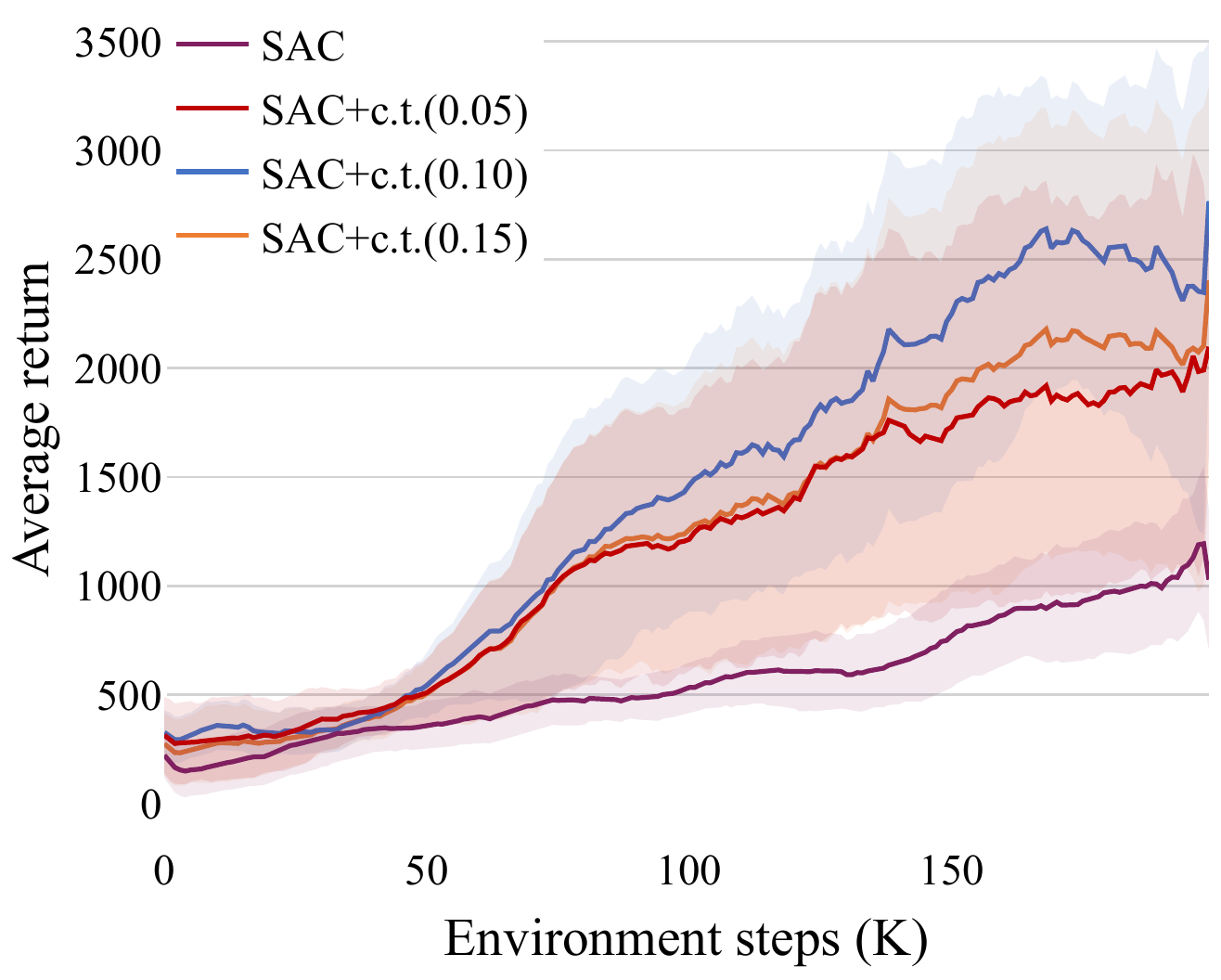}
  \caption{Ant}
  \label{fig:tor_ant}
\end{subfigure}%
\vspace{-5pt}
\caption{Evaluation curves of SAC+c.t. with different tolerance values. SAC+c.t. with different tolerances has consistent performance and outperforms SAC.}
\vspace{-15pt}
\label{fig:tor}
\end{figure*}

\textbf{Environments.} We evaluate the performance of off-policy methods training with \textit{continuous transitions} on four complex continuous control environments from MuJoCo~\cite{todorov2012mujoco}, i.e., Ant, Cheetah, Walker, and Hopper. The observation information of these tasks is in the form of feature~(e.g., velocity and position) and the agent is required to choose an action in continuous space. These tasks are difficult to solve for their high-dimensional states and action spaces.

\textbf{Baselines.} In our experiments, we train the state-of-the-art~(SOTA) actor-critic methods for continuous control tasks, i.e. TD3~\cite{fujimoto2018addressing} and SAC~\cite{haarnoja2018soft2}, with \textit{continuous transition}. We denote SAC+c.t. / TD3+c.t. as SAC / TD3 training with \textit{continuous transitions}, respectively. Besides the comparison with original SAC and TD3, we also compare our methods with the recent SOTA model-based / model-free algorithms. Specifically, following the same experiment settings in SUNRISE~\cite{lee2020sunrise}, we compare our methods with a combination of TRPO~\cite{schulman2015trust} and ensembles of dynamics models (i.e., METRPO~\cite{kurutach2018model}), an advanced model-based~(MB) RL method based on ensembles of dynamics models (i.e., PETS~\cite{chua2018deep}), a SOTA MBRL method (i.e., POPLIN-P~\cite{wang2019exploring}), a variant of POPLIN-P by injecting noise to the action space (i.e., POPLIN-A~\cite{wang2019exploring}),  and a combination of ensemble policies and upper confidence bound exploration strategy (i.e., SUNRISE~\cite{lee2020sunrise}). In contrast to these methods, our method improves the sample efficiency only on the level of data, and neither requires any modification to the original algorithms.

\textbf{Training details.} The hyperparameters for \textit{continuous transitions} across different settings the same. The baselines methods are \ijf{trained} by using the same setting as the publicly repository (https://github.com/vitchyr/rlkit). As for SAC, we adopt the version in~\cite{haarnoja2018soft2}, which tunes the policy entropy automatically. The complete algorithm of the combination of the actor-critic method and \textit{continuous transition} is depicted in Alg.~\ref{alg:ctf}. For simplicity, all network structures in our experiments have the same hidden layers and hidden units. Following the experiment setups in SUNRISE, the curves and results are calculated after 200K timesteps. For better visualization and comparison, all of the curves are smoothed with a window of size five. Each experiment is conducted four times with the mean and the standard deviation reported. The detailed hyperparameters used in our experiments are presented in Tab.~\ref{tab:details}. For those parameters of TD3 that are different from SAC, we mark them with (TD3) in the table.

\begin{table}[t]
\centering 
\vspace{-15pt}
\caption{Values of hyperparameters used in this paper.}
\label{tab:details}
\begin{tabular}{ll|ll}
\toprule
Hyperparameter & Value & Hyperparameter & Value \\ \midrule
\#training frames & 2e5 & \#evaluation interval & 1e3 \\
    Non-linearity & ReLU & Backbone & MLP  \\
   Hidden layers &  2      &     Optimizer           & Adam      \\
    \ijf{Tolerance} $\eta$ & 0.1 & Target update freq & 1 \\
    Hidden units &   256         &  Learning rate     & 3e-4         \\ 
    Batch size &   256   &  Discount $\gamma$ & 0.99 \\
    Initial $\beta$ & 1 & Polyak factor $\tau$ & 0.005 \\
    Replay buffer size           & 1e6      &               \#evaluation episodes &  10 \\
    Actor update freq~(TD3) & 2 & Learning rate~(TD3) & 1e-3 \\
    \bottomrule
\end{tabular}
\vspace{-15pt}
\end{table}

\subsection{MuJoCo Results}
\label{sec:gym}
In this part, we present the performance of all baselines on four continuous control tasks from MuJoCo~\cite{todorov2012mujoco}, i.e., Ant, Walker, Cheetah as well as Hopper. The evaluation results are presented in Tab.~\ref{tab:baselines}. The curves of the evaluation return along the training process are plotted in Fig.~\ref{fig:sac} and Fig.~\ref{fig:td3} for SAC and TD3 respectively.

\begin{figure*}[t]
\centering
\begin{subfigure}{.245\textwidth}
  \centering
  \includegraphics[width=\linewidth]{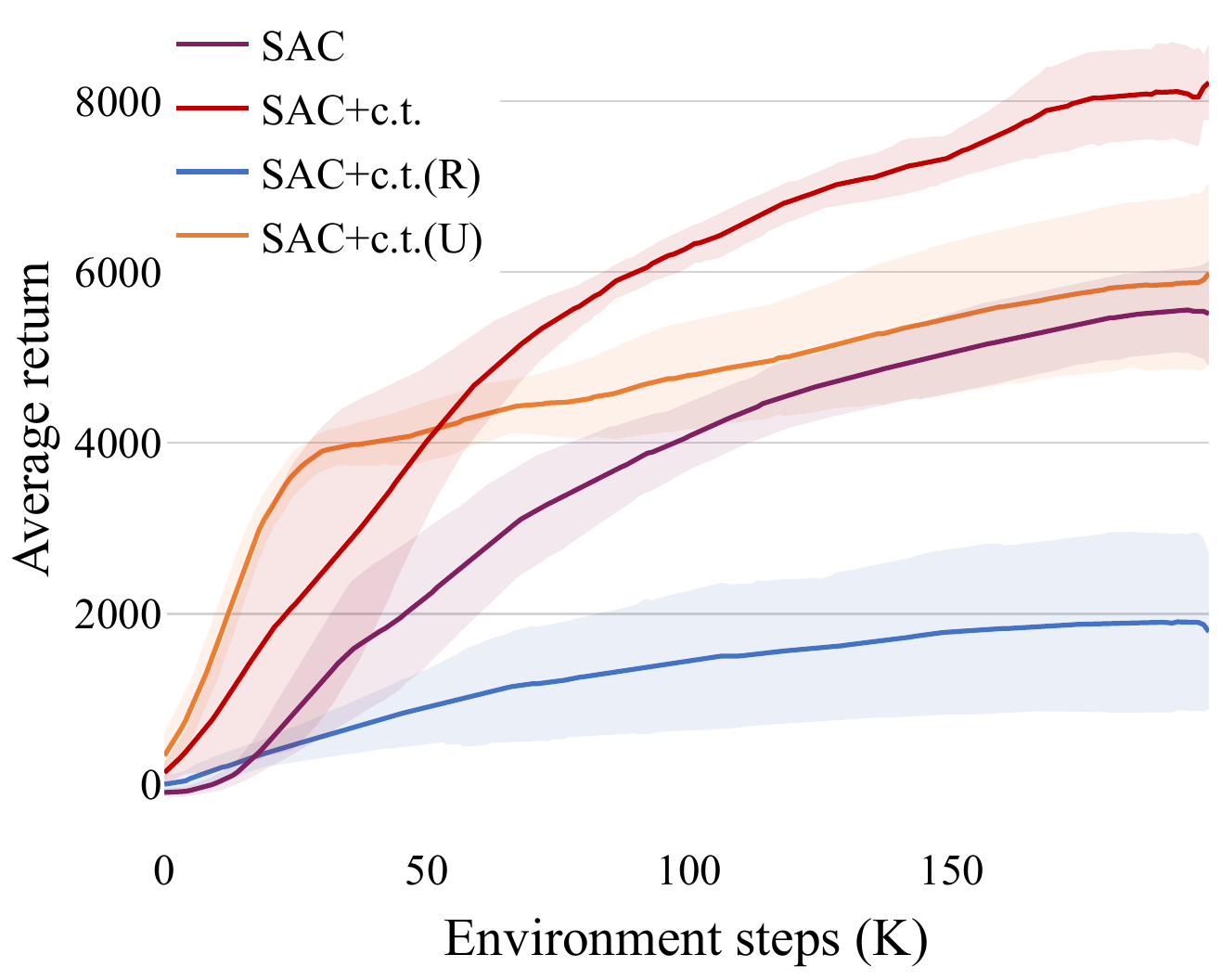}
  \caption{Cheetah}
  \label{fig:manner_cheetah}
\end{subfigure}
\begin{subfigure}{.245\textwidth}
  \centering
  \includegraphics[width=\linewidth]{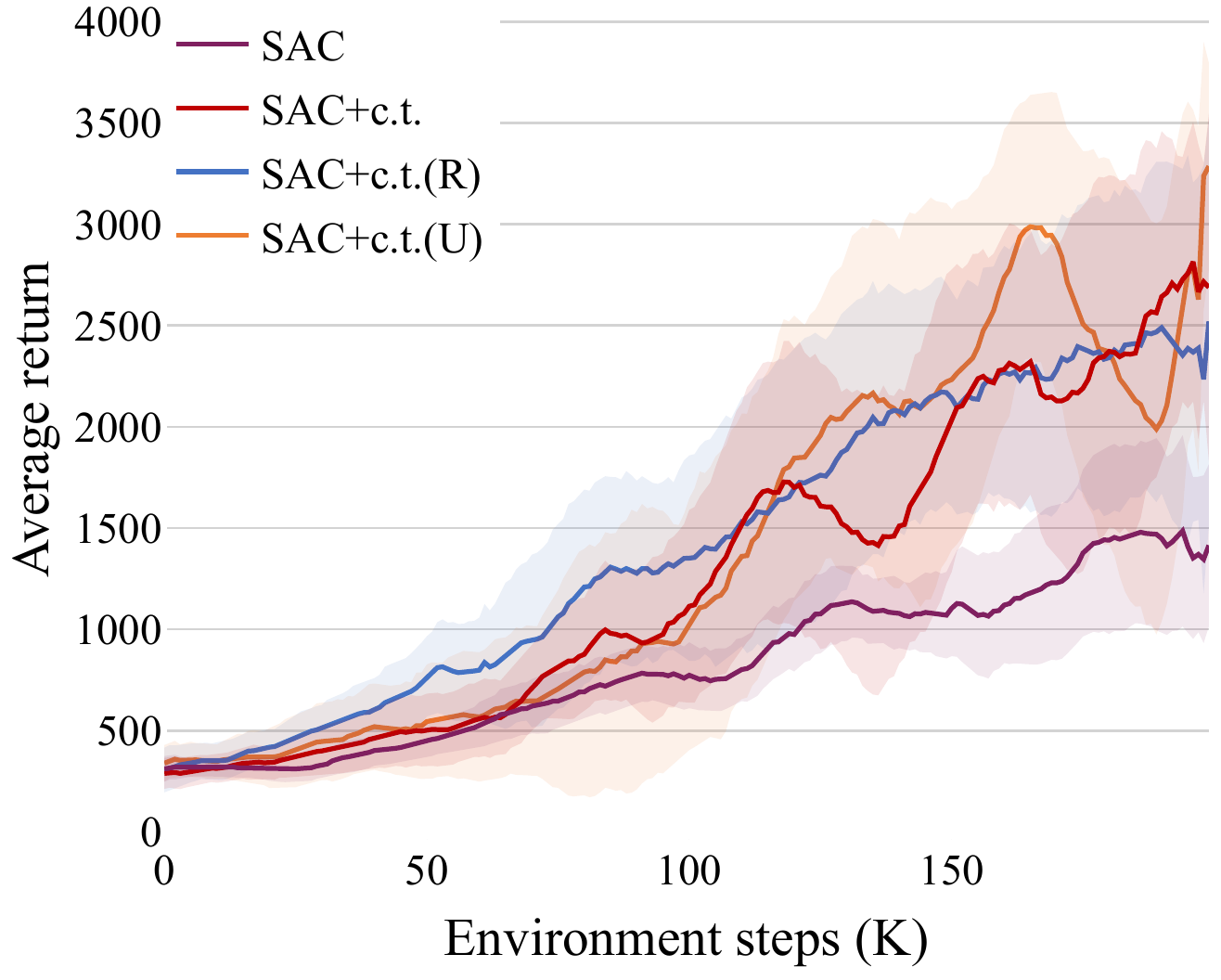}
  \caption{Walker}
  \label{fig:manner_walker}
\end{subfigure}
\begin{subfigure}{.245\textwidth}
  \centering
  \includegraphics[width=\linewidth]{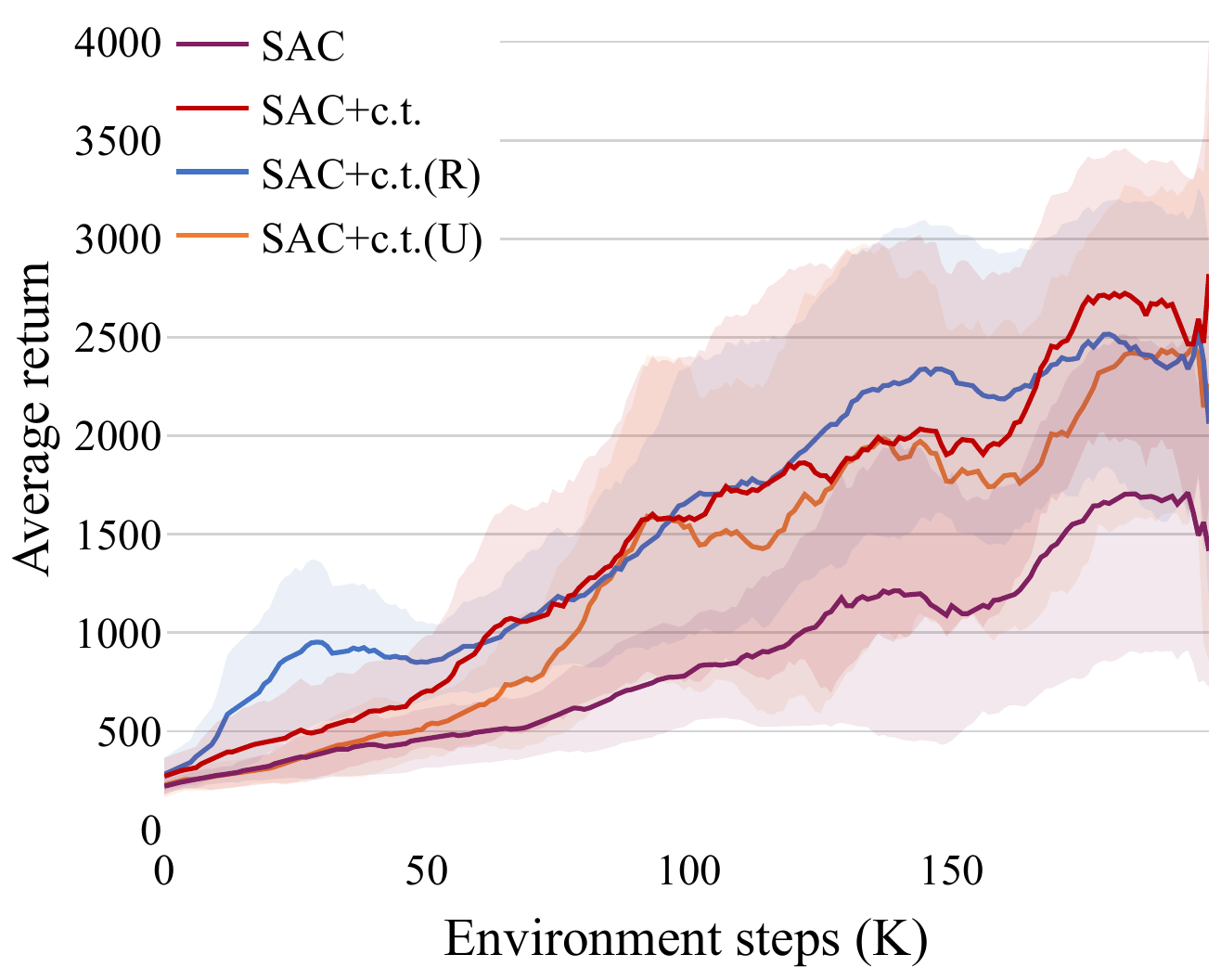}
  \caption{Hopper}
  \label{fig:manner_hopper}
\end{subfigure}
\begin{subfigure}{.245\textwidth}
  \centering
  \includegraphics[width=\linewidth]{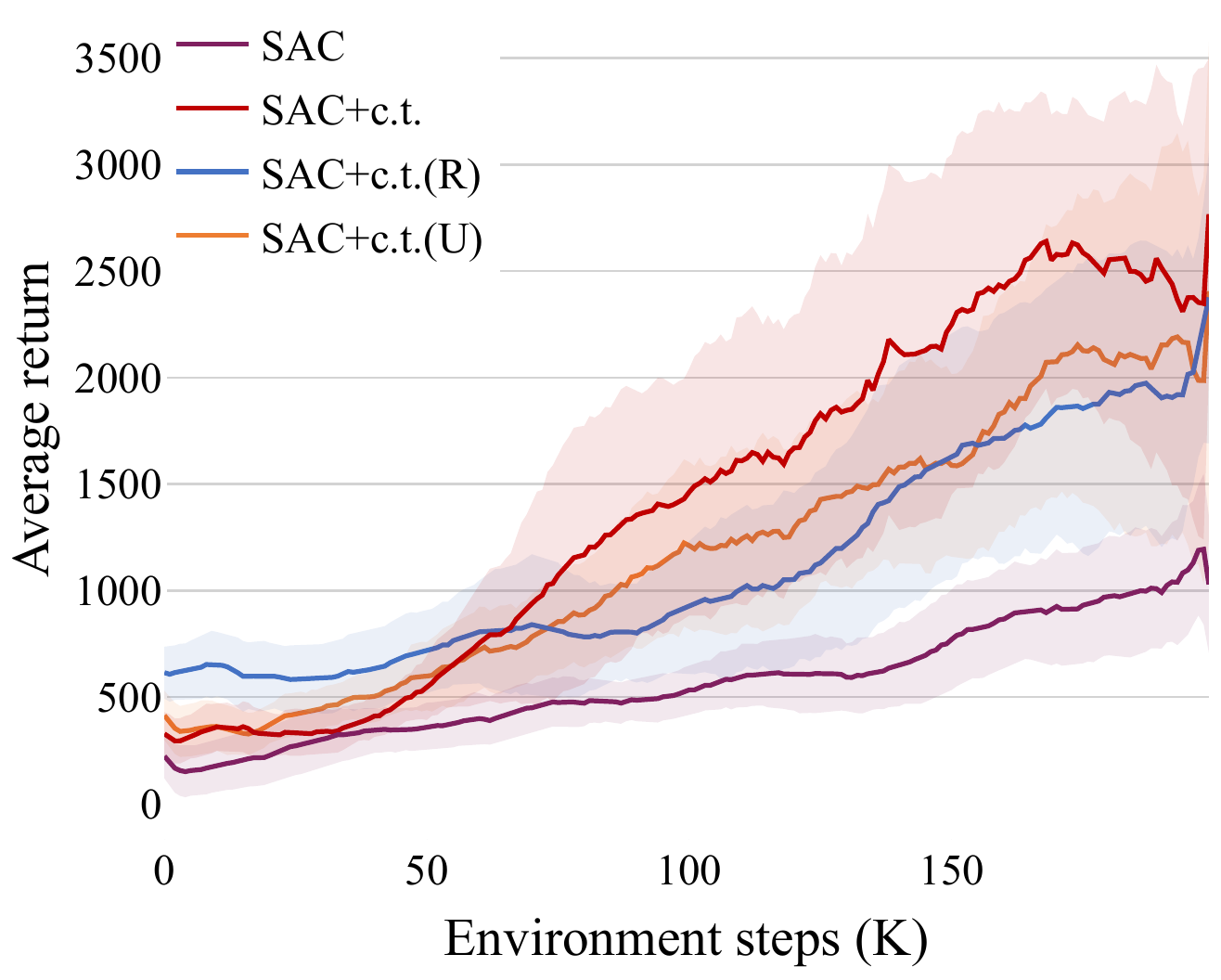}
  \caption{Ant}
  \label{fig:manner_ant}
\end{subfigure}%
\vspace{-5pt}
\caption{Evaluation curves of SAC+c.t. with different interpolation strategies, where +(R) for constructing transitions with random pairs of transitions and +(U) for sampling interpolation ratio from uniform distribution.}
\label{fig:manner}
\vspace{-15pt}
\end{figure*}

\begin{figure}[t]
\centering
\begin{subfigure}{.23\textwidth}
  \centering
  \includegraphics[width=\linewidth]{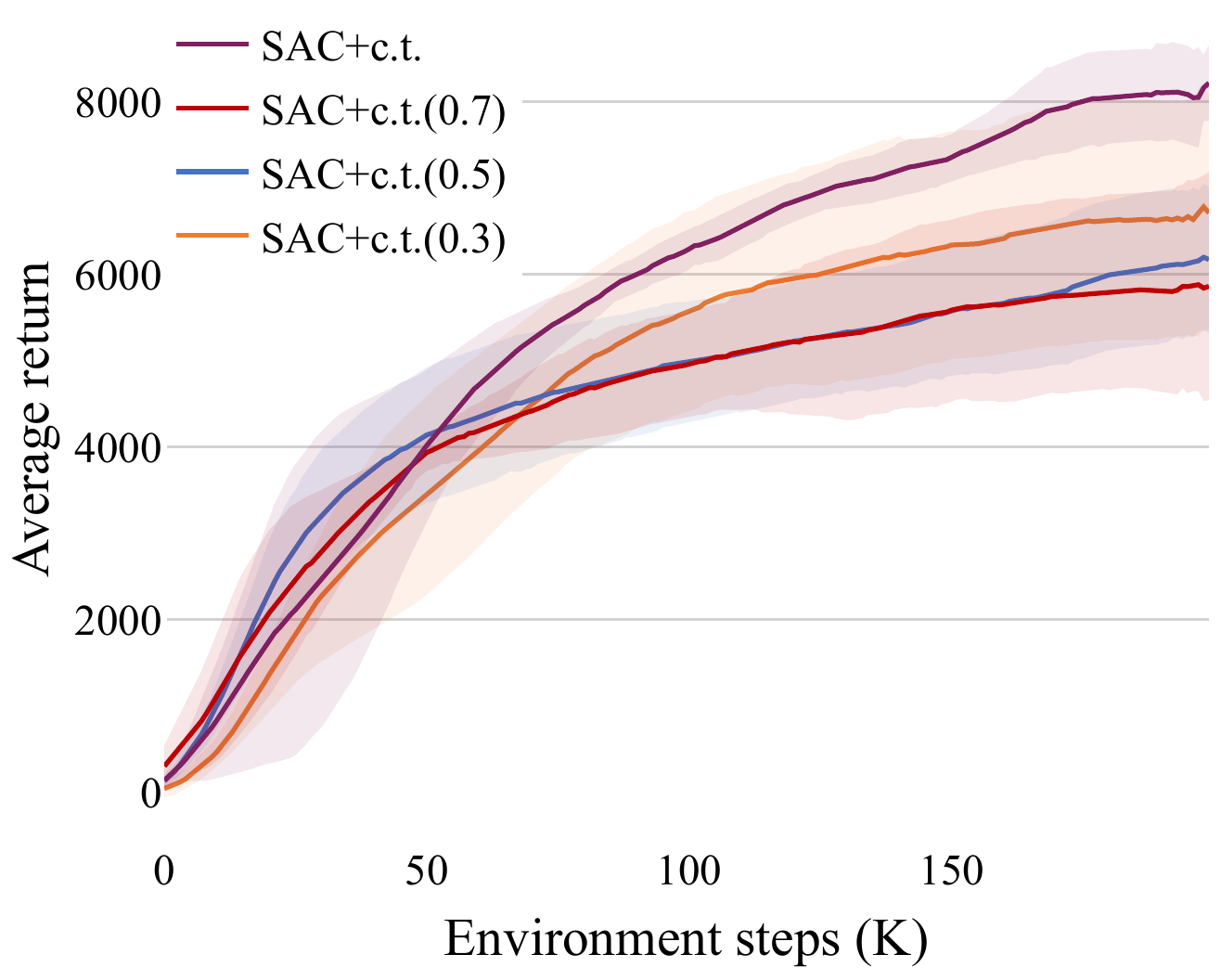}
  \caption{Cheetah}
  \label{fig:cheetah}
\end{subfigure}
\begin{subfigure}{.23\textwidth}
  \centering
  \includegraphics[width=\linewidth]{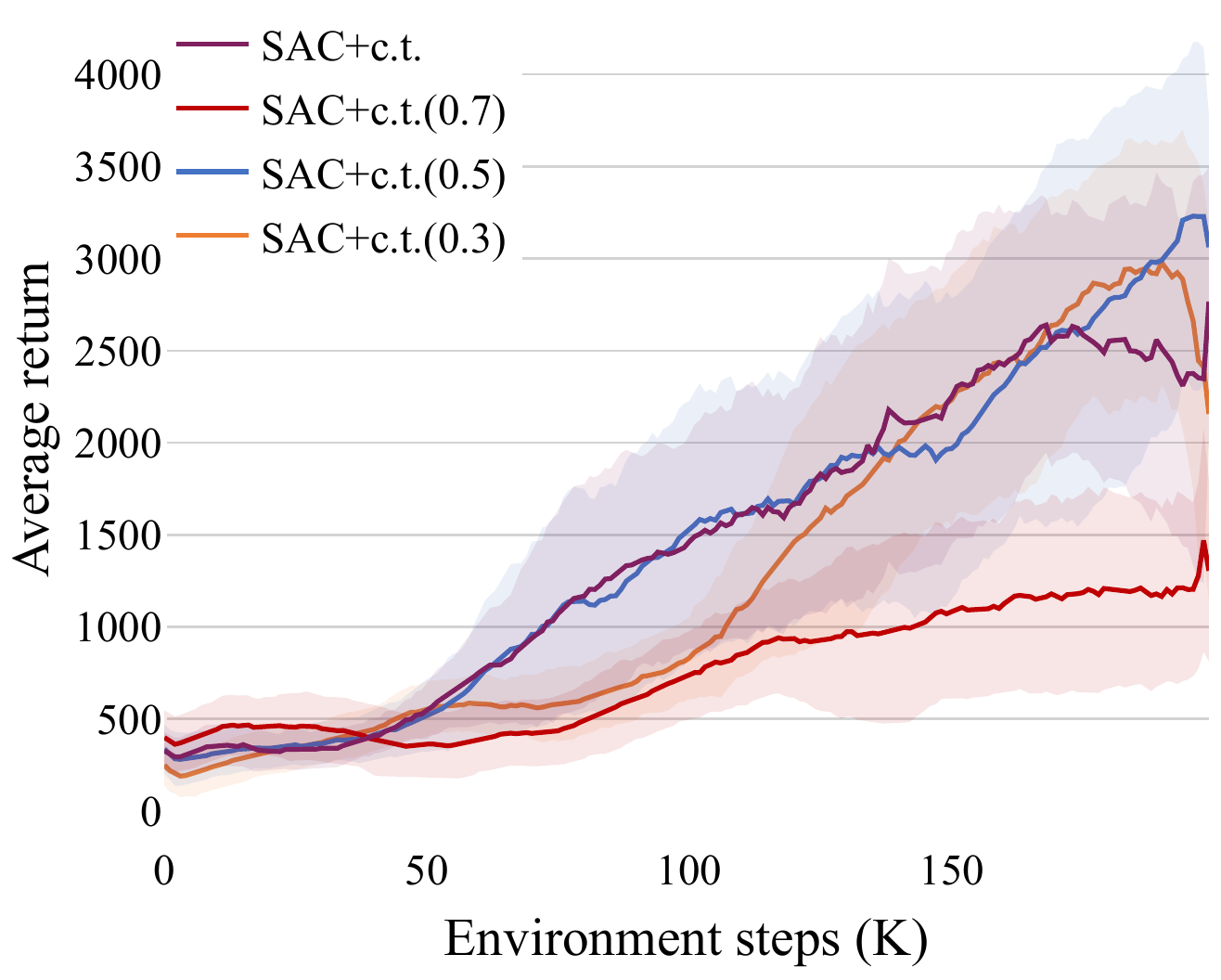}
  \caption{Ant}
  \label{fig:ant}
\end{subfigure}%
\vspace{-5pt}
\caption{Evaluation curves of SAC+c.t. with different fixed beta values in [0.3, 0.5, 0.7]. Auto-tuning SAC+c.t. achieves superior performance on all tasks.}
\vspace{-15pt}
\label{fig:mannual}
\end{figure}

As shown in Tab.~\ref{tab:baselines}, SAC+c.t. outperforms all of the baselines by a clear margin across all tasks. Especially for the performance on Cheetah and Walker, SAC+c.t. consistently makes a breakthrough, i.e., exceeding 8000 and 2500 on average. TD3+c.t. also gains a considerable improvement, compared to TD3 on all tasks. Since that TD3 is unstable on all tasks and its performances with some random seeds are not improved during the whole training process,  there exists a very large standard deviation. This problem is also observed in TD3+c.t.. However, in those experiments that TD3 does not collapse, TD3+c.t. can further improve the sample efficiency resulting in an overall improvement. From the evaluation curves in Fig.~\ref{fig:sac} and Fig.~\ref{fig:td3}, we can observe that the performances of both SAC+c.t. and TD3+c.t. improve at the very early stage of the training process, which implies that \textit{continuous transition} is effective even with a small amount of collected data. In comparison with all baselines, our introduced \textit{continuous transition} is more compute-efficient since it does not require any ensemble of dynamics, any bootstrapping policies and planning during the evaluation phase.

\subsection{Ablation Study}
In this section, we ablate the effects of different values of the tolerance used in the discriminator, and different strategy of interpolation such as interpolation with random pairs of transitions, uniformly distributed interpolation ratio as well as manually determining the temperature of the beta distribution. In this section, we only conduct experiments with SAC and SAC+c.t. since they are more stable.

\label{sec:ablation}

\textbf{Effects of different values of the tolerance.} In the proposed continuous transition framework, the most critical hyperparameter is the tolerance $m$. To evaluate whether our methods are sensitive to $m$, we conduct experiments with different values of $m$. The evaluation curves are shown in Fig.~\ref{fig:tor}, where the values of $m$ are 0.05, 0.1 and 0.15 respectively. From the results, one can find that different values of tolerance have little difference to SAC+c.t, and all of these experiments obtain clear improvements against the original SAC on different tasks. This means that it is not necessary for users to pay many efforts on tuning the tolerance to achieve the improvement. In the experiment of Hopper, we observe that different tolerance resulting in the same results. This is because on this task, $m$ is optimized to the maximum~(i.e., 1) given different tolerance, therefore the curves have collided.

\textbf{Different interpolation strategies.} In this part, we design a set of experiments to demonstrate i) the effectiveness of constructing \textit{continuous transitions} using \ijf{consecutive} transitions, compared to random pairs of transitions, ii) the performance of uniformly sample interpolation ratio, and iii) the benefit of auto-tuning the temperature. To show the effectiveness of choosing the \ijf{consecutive} transition, we construct \textit{continuous transitions} by linearly interpolating random pairs of transitions, denoted as SAC+c.t.(R). As shown by the blue curves in Fig.~\ref{fig:manner}, the performance of SAC+c.t.(R) degrades on most of the tasks. Especially for Cheetah, the performance of SAC+c.t.(R) improves much more slowly than the original SAC. And as for the strategy of choosing the ratio of interpolation, we conduct an experiment that interpolation ratio is a sample from uniform [0, 1] distribution. According to the orange curves shown in Fig.~\ref{fig:manner}, the performances of the controlled experiments are also decreased. However, training with transitions interpolated with a uniform sampling ratio still outperforms the original SAC. This might be because the linear regularization brought by the MixUp operation is helpful to learn a more generalized value approximator~\cite{zhang2017mixup}. Thus, the agent can better estimate bootstrapping value during training and further generalize to out-of-manifold states during testing. To understand the benefit of automatically tuning $\beta$, we conduct a set of experiments with different $\beta \in \{0.3, 0.5, 0.7\}$. As depicted in Fig.~\ref{fig:mannual}, the curves w.r.t. different $\beta$ seem to vary, and automatically tuning $\beta$ can achieve relatively good performance.

\section{CONCLUSIONS}

In this paper, we exploited the potential of the intermediate information along the trajectory by constructing \textit{continuous transition} through linearly interpolating the \ijf{consecutive} discrete transitions. In order to synthesize more authentic transitions, we introduced an energy-based discriminator to auto-tuning the temperature of the beta distribution. Experimental results demonstrated our proposed methods achieve significant improvements on sample efficiency against the compared methods. Considering that our methods are testified in simulated environments, we will extend our work to real robotic control tasks in the future. 





\section*{ACKNOWLEDGMENT}
This work was supported in part by National Natural Science Foundation of China (NSFC) under Grant No.U19A2073, No.61976233, No.U181146, No.61836012, No.62006253, Guangdong Province Basic and Applied Basic Research (Regional Joint Fund-Key) Grant 2019B1515120039, Fundamental Research Funds for the Central Universities (No.19lgpy228), Shenzhen Basic Research Project (No.JCYJ20190807154211365), Zhijiang Lab's Open Fund (2020AA3AB14) and CSIG Young Fellow Support Fund, the State Key Development Program under Grant 2018YFC0830103, Guangdong Natural Science Foundation under Grant 2017A030312006, and DARPA XAI project N66001-17-2-4029.





{
\small
\bibliography{IEEEexample}
\bibliographystyle{IEEEtran}
}

\end{document}